\documentclass[10pt,twocolumn,letterpaper]{article}

\usepackage{iccv}
\usepackage{times}
\usepackage{epsfig}
\usepackage{graphicx}
\usepackage{amsmath}
\usepackage{amssymb}
\usepackage{comment}

\newcommand{\model}{SS-SFDA}

\newcommand{\mc}[1]{\mathcal{#1}}

\makeatletter
\newcommand\footnoteref[1]{\protected@xdef\@thefnmark{\ref{#1}}\@footnotemark}
\makeatother

\newcommand{\shorteq}{%
  \settowidth{\@tempdima}{-}% Width of hyphen
  \resizebox{\@tempdima}{\height}{=}%
}

\usepackage{amssymb,fge}

\usepackage{amsmath, amssymb}
\usepackage{subcaption}
\usepackage[utf8]{inputenc}
\usepackage[T1]{fontenc}
\usepackage[linesnumbered,ruled,vlined]{algorithm2e}
\usepackage[font=small]{caption} % This causes problems while compiling to latex    
\usepackage{array}
\usepackage{graphicx}
\usepackage{amsfonts}
\usepackage{enumitem}
\usepackage{soul}
\usepackage{hhline}
\usepackage{multirow, makecell}
\usepackage{float}
\usepackage{booktabs}

\usepackage{amsthm}
\usepackage{color}
\usepackage{transparent}
\usepackage{footmisc}
\usepackage{setspace}
\usepackage{textcomp}
\usepackage{mathtools}

\theoremstyle{plain}

\mathchardef\mhyphen="2D

\usepackage[pagebackref=true,breaklinks=true,letterpaper=true,colorlinks,bookmarks=false]{hyperref}

\iccvfinalcopy % *** Uncomment this line for the final submission

\def\iccvPaperID{9188} % *** Enter the ICCV Paper ID here

\ificcvfinal\fi

\begin{document}

%%%%%%%%% TITLE
% \title{A Self-Training Algorithm for Unsupervised Road Segmentation in Hazardous Driving Environments}
\title{\model~: Self-Supervised Source-Free Domain Adaptation for Road Segmentation in Hazardous Environments}
\author{
    Divya Kothandaraman,
    Rohan Chandra,
    Dinesh Manocha \\
    University of Maryland, College Park \\
    {\small Tech Report, Code, and Video at  \url{https://gamma.umd.edu/weatherSAfE/}}
}
    
% \affiliations{
%    University of Maryland, College Park \\
%    {\small Tech Report, Code, and Video at %\url{https://gamma.umd.edu/bomuda}}
% }

\maketitle
\ificcvfinal\thispagestyle{empty}\fi

\begin{abstract}

We present a novel approach for unsupervised road segmentation in adverse weather conditions such as rain or fog. % that results in low visibility and hazardous driving conditions (\textit{e.g.} wet patches or potholes caused by rain). 
This includes a new algorithm for source-free domain adaptation (SFDA) using self-supervised learning.  Moreover, our approach uses several techniques to address various challenges in SFDA and improve performance, including online generation of pseudo-labels and self-attention as well as use of curriculum learning, entropy minimization and model distillation.
%that address multiple issues in road segmentation and SFDA. 
% First, we address the problems of saturation and contextual loss in road segmentation via online generation of pseudo-labels and self-attention respectively. Additionally, we use curriculum learning and entropy minimization to bridge the domain gap issue in domain adaptation.  
% We show exhaustive ablation experiments demonstrating the benefits of self-attention, online training, curriculum learning, and entropy minimization. 
%We have evaluated the performance on $6$ datasets corresponding to real and synthetic adverse weather conditions. Our method outperforms all prior works on unsupervised road segmentation and SFDA by at least $10.26$\%, and has faster convergence rates. Moreover, our self-supervised  algorithm exhibits similar accuracy performance in terms of mIOU score as compared to prior supervised methods.
We have evaluated the performance on $6$ datasets corresponding to real and synthetic adverse weather conditions. Our method outperforms all prior works on unsupervised road segmentation and SFDA by at least $10.26$\%, and improves the training time by $18-180 \times$. Moreover, our self-supervised  algorithm exhibits similar accuracy performance in terms of mIOU score as compared to prior supervised methods.
%Overall, our mIoU score is $88-96\%$ of the supervised baseline. Furthermore, 
%We improve the training time over prior SFDA approaches by $18-180 \times$. Moreover, our improvement in terms of mIoU over prior SFDA approaches is at least $10.26$\% on real-world adverse weather data. 

\end{abstract}
\section{Introduction}
\label{sec:intro}

Research in autonomous driving continues to advance in terms of improving the perception capabilities of self-driving cars for greater safety. This includes detecting both static and dynamic obstacles such as pedestrians~\cite{dollar2011pedestrian}, tracking and predicting the trajectories of other vehicles~\cite{chandra2019traphic,chandra2020forecasting, chandra2019robusttp, chandra2020roadtrack}, and scene segmentation~\cite{fu2019dual,zhao2017pyramid}. Immense progress along these lines has led to the deployment of level $2$ and almost level $3$ autonomous vehicles (AVs) in urban traffic environments~\cite{cusumano2020self}. However, these advances in perception technology have been primarily designed to work well in safe and clear weather conditions. Driving in adverse weather and lighting conditions such as snow, rain, or fog is challenging not just for autonomous vehicles but even for humans. These conditions result in a degradation in accuracy of perception techniques including road segmentation~\cite{sakaridis2018model,sakaridis2019guided}. Consequently, AVs are unable to distinguish  drivable regions of the road from the non-driveable region (which may be affected by snow, rain, or fog), thereby increasing the likelihood of road accidents~ \cite{mueller2012driving}. In this paper, we address the problem of road segmentation in adverse weather conditions.

% Driving in adverse weather and lighting conditions is challenging not just for autonomous vehicles, but even for humans. Accident prevention in these unfavourable environmental conditions necessitates that the autonomous vehicle or human driver have a clear semantic understanding of the scene. Road segmentation is important for path planning, trajectory prediction and drivable region detection. Recent advances in deep learning have been instrumental in fostering self-driving research and have improved performance of road segmentation algorithms. The inherent nature of the task of road segmentation drives the network to focus on drivable areas of the scene, and limit attention from regions that are not suitable for driving. This fosters the development of methods that can be applied across diverse challenging driving conditions as opposed to individual solutions that rely on deraining, dehazing, and intensity modification methods to develop architectures for driving in rain, haze, and under low light conditions respectively. 
\begin{figure}[t]
    \centering
    \includegraphics[width=\columnwidth]{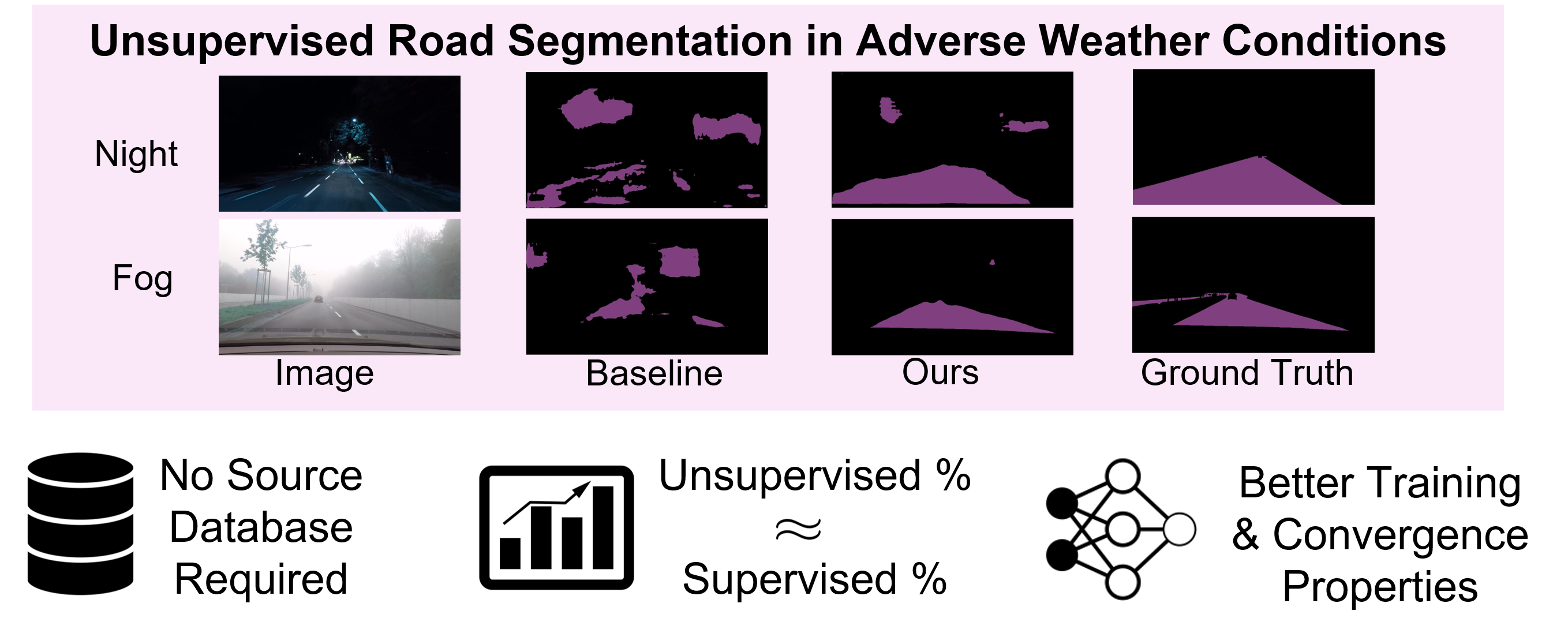}
    \caption{We highlight the results generated by SS-SFDA on night \cite{sakaridis2019guided} and fog benchmarks \cite{sakaridis2018model}, compared to the baseline source model pre-trained on clear weather CityScapes. The purple regions (right) denote the segmented road pixels. The overall accuracy of our self-supervised algorithm in terms of mIoU  is ($88-96\%$) of supervised methods.}% and we observe considerable improvement over prior approaches for road segmentation and SFDA.}
    \label{fig: coverpic}
    \vspace{-12pt}
\end{figure}

The road segmentation ~\cite{fan2020sne,sun2019reverse} problem corresponds to identifying the pixels in an RGB image or video that belong to the `road' class. While general models designed for semantic segmentation in computer vision can be directly used for road segmentation, they suffer from the inability to capture semantic relationships between different objects due to the lack of unique labels for each class. The use of self-attention techniques ~\cite{fu2019dual,zhang2019self} can mitigate this issue by capturing long-range dependencies. 

%Current work in supervised road segmentation in clear weather suffers from loss of context due to the unavailability of semantic labels for all classes in the scene as many objects such as cars, pedestrians, etc. are categorized as ``non-road''. The loss of context adversely affects the capability of the current road segmentation to adapt to adverse weather conditions. The usage of self-attention in road segmentation~\rohan{cite Safe arxiv paper} has been shown to mitigate the degradation in performance due to loss of context.

However, one major challenge in road segmentation in adverse weather is the lack of ground-truth annotations for road pixels. A common approach in deep learning for handling lack of training data is domain adaptation (DA)~\cite{hoffman2018cycada,vu2019advent}. However, DA-based methods assume access to source datasets (clear weather dataset in our context) at all times which can be prohibitive in terms of storage, memory, data corruption and privacy concerns. Recently, \cite{kim2020domain, nelakurthi2018source} have proposed source-free domain adaptation (SFDA) in which deep neural networks (DNNs) do not require access to the source dataset during the adaptation stage; instead, DNNs are pre-trained on a source dataset (clear weather dataset) and the pre-trained model is directly used to adapt to the unlabeled target domain (adverse weather dataset). %However, SFDA methods suffer from the domain gap problem which corresponds to the difference between the source pre-trained model and target domains due to which pre-trained model may not perform well. %such methods are unable to perform as well as those methods that are trained via supervised learning%However, prior work has mainly focused on SFDA% in an unsupervised manner.
% target domain in an for inference on the target domain. 

Current methods for SFDA are used for image classification~\cite{kundu2020universal,hou2020source,kurmi2021domain,yeh2021sofa} and may not work well for semantic segmentation due to the inherent differences between the classification and segmentation tasks. Moreover, many current SFDA methods use GANs to produce a ``copy'' of the original source domain distribution. In addition to being computationally intensive due to the difficulty in training GANs, image generation for segmentation requires GANs to capture contextual information and semantic relationships between multiple objects and the background, which can be complicated in road scenes. As a result, prior SFDA techniques have not been used for road segmentation.
%To the best of our knowledge, this paper marks one of the first approaches towards SFDA road segmentation. 
%Furthermore, SFDA methods inherently suffer from the domain gap problem which corresponds to the difference between the source and target domains due to which such methods are unable to perform as well as those methods that are trained via supervised learning (\textit{i.e} having access to annotated labels for the target domain)

\subsection{Main Contributions}

We present a new approach for road segmentation in adverse weather conditions. Our approach is based on a novel algorithm for SFDA using self-supervised learning. We initialize our model with an auto-encoder baseline network using self-attention to generate a pre-trained model on the clear weather source dataset. Using self-attention improves the overall model by capturing long-range dependencies within the image (Section~\ref{sec:sup}). Our novel contributions include: 
%The main advantages of self-supervised learning in SFDA over the state-of-the-art generative process is that \rohan{mention problems with GAN-based SFDA}. 
% Our main contributions are summarised as follows:
\begin{enumerate}[nosep]
%Components of SS-SFDA:
%Initialization - faster convergence, prior information about road pixels, directly exploited by SS-SFDA instead of data generation by GANs/ prototype metric alignment, etc which works for classification but not segmentation
%2 steps: 
%Step 1 EM - pre-trained initialized model is noisy, so direct self-training with pseudo labels won't help; EM generates enriched pseudo labels
%Step 2 Self training with PL - online generation to tackle saturation
%Entire training procceds with curriculum learning (over multiple minibatches from low entropy to high entropy) - for each minibatch, Step 1 and step 2 are performed
%All three components (CL{EM->PL} helps progressively bridge domain gap between pre-trained model and target domain.
%CL - additionally helps in convergence bevause of progressive training

% \item We present a novel self-supervised SFDA approach called \model. \model~uses the self-attention auto-encoder at initialization resulting in faster training time. Since the pre-trained model is sub-optimal on the target domain, which is detrimental to self-training, \model~proposes an entropy minimization step to generate enriched pseudo labels which is followed by a self-training step. 

\item We present a novel two-step self-supervised SFDA approach called \model. In the first step, our method uses entropy minimization to enrich the noisy pseudo-labels generated by the pre-trained auto-encoder. In the second step, we use a novel self-training method that generates pseudo labels in an \textit{online} manner, as opposed to iterative self-training use by prior methods~(Section \ref{sec:unsup}). We use curriculum learning to implement these two steps.  This results in the following benefits compared to prior GAN-based approaches:

\begin{itemize}[noitemsep]
    \item SS-SFDA directly exploits the pre-trained model and trains via curriculum learning to progressively bridge the domain gap between the pre-trained source domain and target domain and achieve faster training times.
    % in addition to alleviating the issues of noisy predictions and improving training time. % ($88\%-96\%$ of supervised models).
    %\item Pre-trained initialization results in faster training time ($18-180 \times$)
    
    \item Our online self-training scheme overcomes the saturation issues.
    % \item 
\end{itemize}

% WHAT IS NEW AND UNIQUE ABBOUT YOUR METHOD THAT MAKES IT WORK FOR ROAD SEGMENTATION, WHILE THE PRIOR SFDA METHODS FAIL. THIS . %and learns robust feature representations on the target dataset via curriculum %and generates pseudo-labels in an online manner resulting in faster convergence and bridging the domain gap.
% We show that SAfE is comparable to complex state-of-the-art methods (Section \ref{sec:sup}). 

\item For heterogeneous adverse weather datasets, we propose a method that extends \model~by leveraging a few labeled images from the target domain to improve the accuracy using model distillation (Section \ref{sec:fewimage}).
\end{enumerate}

%We propose a new approach for SFDA via self-supervised learning. We pre-train the base model using self-attention that captures the semantic information pertaining to the `road' class. We also bridge the domain gap with curriculum learning by minimizing an entropy-based loss function. 
We have evaluated our approach on $6$ datasets corresponding to real and synthetic adverse weather conditions. Overall, our mIoU score is $88-96\%$ of prior supervised methods. We also improve the training time over prior SFDA approaches by $18-180 \times$. Finally, our improvement in terms of mIoU over the best SFDA approach is $10.26$\% on real adverse weather data. 

\section{Related Work}

We discuss recent work related to road segmentation, domain adaptation and source-free domain adaptation, and self-supervised learning.

\subsection{Road Segmentation}
Research in deep learning for semantic segmentation~\cite{long2015fully, yu2017dilated, chen2017deeplab,chen2017rethinking, zhao2017pyramid, fu2019dual, takikawa2019gated} has paved the way for segmentation in urban traffic scenes like CityScapes~\cite{cordts2016cityscapes}. 
% Starting with fully convolutional networks~\cite{long2015fully}, the development of Dilated Residual Networks \cite{yu2017dilated}, DeepLab~\cite{chen2017deeplab,chen2017rethinking}, PSPNet~\cite{zhao2017pyramid}, Dual Attention Nets~\cite{fu2019dual}, GatedSCNN~\cite{takikawa2019gated} and many more architectures~\cite{hao2020brief} has led to increasing accuracies on urban scene datasets like CityScapes~\cite{cordts2016cityscapes}.
These methods have been extended for supervised road segmentation~\cite{wang2017embedding,zohourian2018superpixel,fan2020sne,sun2019reverse}. Our approach based on self-supervised learning is  complimentary to these methods.

% However, these models have an overhead in terms of computation and complexity. In this paper, we show how the incorporation of self-attention within "off-the-shelf" segmentation architectures to explicitly focus on roads and capture long range dependencies can lead to performance that is comparable to the state-of-the-art in road segmentation.

\subsection{Domain Adaptation and Source Free Domain Adaptation}
\label{sec:relatedwork_da}

Traditional domain adaptation \cite{hoffman2016fcns, hoffman2018cycada, sankaranarayanan2017unsupervised, vu2019dada, tsai2018learning, chen2017no} methods have achieved remarkable success in adapting models from one domain to another for clear weather conditions. However, these methods  need access to the source data. Many domain specific solutions  have been proposed for  adverse weather conditions, including specific solutions  for driving in rain, fog, etc.~\cite{porav2019can,mueller2012driving, sakaridis2018model,dai2020curriculum,sakaridis2018semantic, pizzati2020domain}  In contrast, we propose a generic method that neither relies on specific details from each domain, nor requires access to source data during the adaptation stage.

\begin{figure*}[t]
     \centering
     \includegraphics[width=\linewidth]{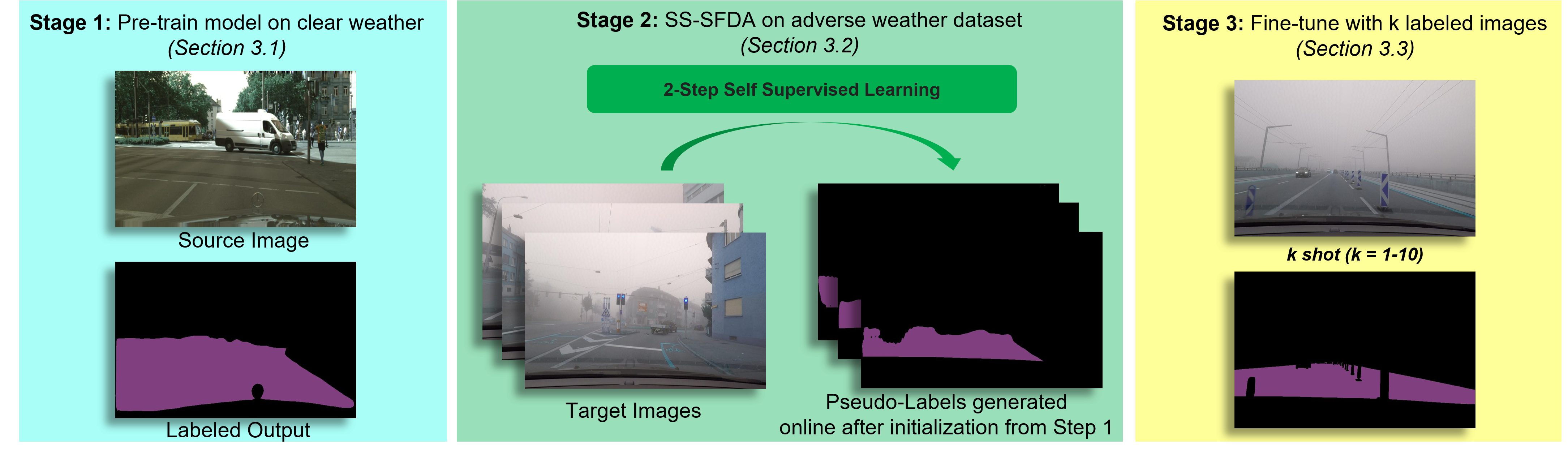}
     \caption{\textbf{Our Approach:} In stage $1$, our model is pre-trained on a clear weather source dataset. In stage $2$, our model is initialized with the pre-trained model from stage $1$ and trained using our self-supervised algorithm, \model, on the unlabeled adverse weather dataset. For heterogeneous weather datasets, we perform additional refinement steps based on model distillation (stage $3$).}
     \label{fig: overview}
     \vspace{-15pt}
 \end{figure*}

%Driving in hazardous weather conditions like snow, fog and rain \cite{valada2017adapnet}, and on poorly maintained roads (potholes, debris, puddles) \cite{yarram2018city} still remains a challenge. %While there are techniques for object detection \cite{xu2019wasserstein, rodriguez2019domain} in rain and fog, these methods can't segment out roads. In the context of segmentation, 
%Prior literature has seen the development of specific solutions \cite{porav2019can,mueller2012driving, sakaridis2018model,dai2020curriculum,sakaridis2018semantic, pizzati2020domain} for driving in rain and fog, which use domain bridges, deraining and dehazing priors, etc which may be hard to obtain. Additionally, these methods require access to the source dataset which may be unavailable. In contrast, we propose a generic architecture that does neither relies on specific details from each domain, nor requires access to source data during the adaptation stage. 

In source-image free domain adaptation (SFDA), a deep neural network (DNN) pre-trained on a source dataset is required to directly make predictions on the target domain dataset in an unsupervised manner. This approach has been primarily used for image classification tasks. In SFDA, generative approaches~\cite{kundu2020universal,hou2020source,kurmi2021domain,yeh2021sofa,li2020model} are used to either emulate the source data by using the feature representations of the pre-trained model or create a negative source dataset during the pre-training stage. Non-generative approaches~\cite{kim2020domain,yang2020unsupervised,liang2020we,ishii2021source} rely on computing adaptive class specific prototypes, and progressively learn on the target images. While generative approaches work well for classification tasks, they have not been effective for urban scene segmentation. This is because segmentation, being a pixel level task, requires the network to encode context, inter-class semantic relations, structure, and intricate boundaries~\cite{ding2018context,zhang2018context,zhou2019context}. In addition, training a generator for a complex segmentation task can lead to   memory and computational overheads, and thereby adds to the difficulty of training GANs. While non-generative classification methods are easier to adapt for segmentation, the computation of prototypes  results in similar issues due to the inherent differences between classification and segmentation tasks. Bateson et al.~\cite{bateson2020source} explored SFDA in the context of medical segmentation and minimized entropy by the incorporation of class priors~\cite{bateson2020source}. However, preservation of class priors does not extend well to urban road scenes because the number of road pixels in each image in the dataset can vary quite drastically. In contrast, we present a new method for SFDA using self-supervised learning that overcomes issues related to training and convergence that are freq generative processes. Our method is designed to be complimentary to GAN-based SFDA.

\subsubsection{Self-Supervised Learning}
Self-supervised learning has been used in semi-supervised learning~\cite{li2019learning,cascante2020curriculum} and domain adaptation~\cite{shin2020two}. Most self-supervised learning methods are centered around the ideas of pseudo labeling~\cite{lee2013pseudo,choi2019pseudo,morerio2020generative,zou2018unsupervised,kothandaraman2020bomuda}, entropy minimization~\cite{vu2019advent,grandvalet2005semi}, and curriculum learning~\cite{zhang2017curriculum}.
% For instance, semi supervised learning methods contain labeled images for one part of the dataset
Domain adaptation methods have access to source domain images. In contrast, we propose a novel self-training routine for SFDA and a completely unsupervised problem setting, where we have access to only a pre-trained model, and target domain images.
%\subsubsection{Curriculum learning}
%Cover traditional curriculum learning, intra-target domain adaptation

%\subsection{Few-image}
%Further improve performance with few-image learning
%Knowledge distillation
%Survey on past work, one line about our work

%\section{Source-Image Free Domain Adaptive Road Segmentation via Self-Supervised Learning}
\section{Our Approach}

In this section, we present our approach for source-image free domain adaptive (SFDA) road segmentation based on self-supervised learning. Our approach consists of three main components: %Our method doesn't need access to source data after the initial pre-training step, and is unsupervised in the target domain. Since our model does not attempt to emulate source data by generative approach, it does not incur any additional computational costs or memory overheads. Furthermore, we show that the performance of our approach is comparable to its counterpart supervised learning model. We summarize our approach as follows,

% We present our baseline architecture for road segmentation in Section \ref{sec:sup}, which is used as the backbone for all further experiments. In Section \ref{sec:unsup}, we elucidate the self-training approach for source image free domain adaptive road segmentation, which uses the model described in Section \ref{sec:sup}. In Section \ref{sec:fewimage}, we build on the trained model in Section \ref{sec:unsup} and propose a few labeled-image method that further boosts performance in extremely complicated environments. \divya{summary}

\begin{enumerate}[nosep]
    % \item[] \textbf{Algorithm}
    \item \textbf{Pre-training using the self-attention auto-encoder:} During this stage, we train the self-attention auto-encoder architecture on a clear weather dataset. This generates a model that encapsulates knowledge about road pixels.
    % However, the problem that hinders source-free learning in addition to unavailability of target domain labels is the information gap between the pre-trained model and target domain images due to differences between the source and target domains\cite{wang2018deep}. We propose Steps 2 and 3 for solving this. (Section \ref{sec:sup})
    %However, the problem that hindered source-free learning was the information gap between the pseudo labels and target domain images (any formal term for this gap? prior work?). We propose Steps 2 and 3 for solving this.
    \item \textbf{\model~: A Self-Supervised learning algorithm for SFDA:} During this stage, we initialize the model using the pre-trained model from the previous step and the target domain images. We use a combination of curriculum learning and entropy minimization to bridge the domain gap between the pseudo-labels and the target domain images. We first sort the target domain images in the increasing order of entropy, and create mini-batches of the dataset. The next task is to execute the following steps on each mini-batch to progressively self-train the model:
    %Use Curriculum Learning to sort the target domain images in increasing order of entropy (mention why entropy? is entropy an equivalent form of prediction score?)
    \begin{itemize}[nosep]
        \item Optimize the model with an entropy minimization constraint to bridge the domain gap.
        \item Self-train the model by generating enriched pseudo-labels in an online manner.
    \end{itemize}
    \item \textbf{Few-Image Regularization:} For heterogeneous weather datasets, we use a very small number of labeled images ($5-10$) from the target domain to boost the performance of \model~via model distillation. (Section \ref{sec:fewimage})
      
\end{enumerate}

\subsection{Pre-Training Baselines Using Self-Attention}

\label{sec:sup}
The first step in SFDA is to pre-train a DNN on the source dataset for the task of road segmentation. In our case, the source dataset corresponds to traffic videos with clear weather conditions. While networks developed for semantic segmentation \cite{takikawa2019gated,zhao2017pyramid,chen2017deeplab,chen2017rethinking} can be directly used for road segmentation, there is loss of context i.e. the model is unable to capture relationships between various semantic classes in the images like cars and roads, pedestrians and roads, sky and roads, etc. The loss of such context can lead to local ambiguities in classifying pixels~\cite{ding2018context,zhang2018context,zhou2019context}.

% This loss of context can be attributed to two reasons: (i) in the road segmentation problem, objects like cars and pedestrians and other background classes are not uniquely labeled (all classes other than `road' come under the umbrella class `non-road') (ii) the receptive field of convolution layers is not large enough to capture these relationships.

Self-attention benefits from its capability to capture long-range dependencies between various regions of the image. Thus, using self-attention in road segmentation can allow neural networks to alleviate the degradation in performance due to loss of context. We use a simple autoencoder self-attention architecture that can be combined with any ``off-the-shelf'' segmentation network. We begin by taking an input RGB image $\mathcal{I}$ $\in \mathbb{R}^{w\times h \times 3}$, which is passed through an encoder $E$ to generate feature maps $F_{en}$ ($F_{en}=E(I)$). Next, we apply self-attention $SA$ \cite{zhang2019self} on these features maps to obtain \textit{attention} maps $F_{sa}$ (same dimensions as $F_{en}$). These feature maps encapsulate the semantic relationships between various parts of the image. These feature maps $F_{sa}$ are used to learn the final predictions $\mathcal{P}_{out}$ $\in \mathbb{R}^{w'\times h' \times 1}$, which corresponds to the probability that each pixel is classified as `road'. In the supervised setting where ground-truth labels $\mathcal{Y}$ $\in \mathbb{Z}^{w'\times h' \times 1}$ are available, the network is optimized with a binary cross-entropy loss function,
\begin{equation}
    \mc{L}_{\textrm{CE}} = - \sum_{h,w}  \mc{Y}\log(P_{out}) + (1-\mc{Y})\log(1-P_{out}).
    \label{eq: cross_entropy}
    \end{equation}
\noindent We use this pre-trained model for SFDA on the adverse weather datasets~\cite{sakaridis2018model,sakaridis2019guided,yu2018bdd100k,halder2019physics,tung2017raincouver}.

\subsection{\model~}
\label{sec:unsup}

In this section, we describe our two-step self-supervised learning algorithm for unsupervised source-image free domain adaptive road segmentation in adverse weather conditions. We use the pre-trained model from the previous step. The pseudo-labels generated by the model are noisy leading to a domain gap between the pseudo-labels generated by the pre-trained model (on the source dataset) and the target domain images. Thus, directly self-training using the pseudo labels can hamper the performance of the model. To counter this, we propose an entropy minimization step (Section \ref{sec:ent_min}) which encourages the network to generate more accurate pseudo labels.
%These enriched pseudo labels can then be used to boost performance.

In addition, to bridge the domain gap between the pre-trained model and the target domain, we use curriculum learning~\cite{bengio2009curriculum, hacohen2019power} in which the DNN is allowed to train on samples progressively in their increasing order of entropy of predictions. Given a probability map $P$ denoting the probability that pixels are classified as road pixels, the entropy is computed as $-\Sigma P \times \log(P)$ %, which is a measure of the confidence or probability with with which the network predicts labels for a given image. 
This is because learning from samples with low entropy (low rain, for example) yields better pseudo labels on samples with higher entropy (high rain, for example)~\cite{dai2020curriculum,zhang2017curriculum,zhang2017curriculum}. We create mini-batches of the dataset characterizing the difficulty of the images. For datasets which provide labels on the intensity (light rain vs heavy rain) of the weather condition, mini-batches can be created directly. For other datasets, we sort the images in increasing order of the entropy and then split them into $m$ ($m \sim 4-5$, determined by hyperparameter tuning) equal mini-batches. The model is self-trained on the mini-batches in a sequential manner. For the first mini-batch, the model is initialized with the pre-trained model from Section~\ref{sec:sup}. For subsequent mini-batches, our model is initialized with weights obtained by training the network on the previous mini-batch. For each mini-batch, the network is trained in two stages, as described below:

\subsubsection{Step 1: Bridging the Domain Gap via Entropy Minimization}
\label{sec:ent_min}
% \paragraph{Problem set-up:}
% In the source free unsupervised domain adaptation problem setting, we are provided with a model pre-trained on a clear weather dataset (also called as source dataset). We do not have access to the source dataset after the pre-training step. The task is to learn an optimal feature representation for the adverse weather target dataset. While we have access to the images in the target dataset, the ground truth road segmentation maps are unavailable.
% \paragraph{Method:}
% Since ground-truth annotations are unavailable, it becomes imperative to set up a cost function for training using the pre-trained model and target domain images. Prior information about the feature representations for pixel captured by the pre-trained model can be used adapt to target domain images. One way to do this is to self-train the model using the pre-trained model's predictions (also referred to as pseudo labels) on the target domain images.

% A hard sample Typically, the pseudo labels generated from the easy samples tend to have low noise and vice versa. Moreover, as the training proceeds, the performance of the network on the hard samples also improves. We characterize the `difficulty' of the samples by the entropy of the predictions of the pre-trained source model. 
The pre-trained model from Section~\ref{sec:sup} has a low entropy (\ie high prediction probability or better generalization) on images that are similar (for example, similar geography, light rain, light fog)
to source domain images \cite{sakaridis2019guided} and vice versa. %Conversely, the pre-trained model has a high entropy (or low prediction probability) on images which are very different (for instance, target domain images depicting heavy rain in completely different geographies) from source domain images. 
Thus, initializing the network with these pre-trained weights, followed by training by entropy minimization ~\cite{grandvalet2005semi,saito2019semi} allows the network to generate enriched pseudo-labels. The inputs to the network are images from the target domain. Let the predictions of the network be denoted by $\mathcal{P}_{out}$ $\in \mathbb{R}^{w'\times h' \times 1}$, corresponding to the probability that each pixel is classified as `road'. The cost function for entropy minimization is given by,
    \begin{equation}
        L_{\textrm{EM}} = - \sum_{\forall \textrm{pixels}}P^{h,w} \log P^{h,w},
    \end{equation}
    
\noindent where $P^{h,w}$ is the probability that a pixel belongs to a class `road' at a given location, and $-P^{h,w} \log, P^{h,w}$ is the entropy.

% However, the performance of the pre-trained network on the `easy' samples is not optimal either. Learning high quality pseudo labels requires the network to be able to predict with both (i) high accuracy and (ii) high confidence. High confidence (or high probability) predictions corresponds to low entropy (or low uncertainity). For a binary prediction task, entropy and accuracy are highly correlated. Thus, minimizing the entropy of the predictions can lead to better pseudo labels. 

% These improved pseudo labels can be used to further enhance the performance of the network by self-training.

\subsubsection{Step 2: Online Self-Training Using Enriched Pseudo-Labels}
\label{sec:st_pl}

The network trained in Step 1 generates enhanced pseudo labels with high probability, and is thus a better representative of the target domain than the pre-trained source model from Section~\ref{sec:sup}. These enriched pseudo labels which can be used to self-train the model further to improve performance. A traditional method of self-training using pseudo-labels is iterative~\cite{pan2020unsupervised} in which the network is trained to convergence (validation loss less than a given threshold) over multiple iterations. In each iteration, pseudo labels from the trained model in the previous iteration are used to set up the binary segmentation cost function. We show (Table \ref{tab:sota}) that iterative self training does not lead to any improvement in performance. This is because, the pre-training step, which is imperative for acquiring initial knowledge about road pixels since the problem is unsupervised in the target domain, causes the network to saturate quickly. Hence, we generate the pseudo labels in an online manner, \textit{i.e.} the pseudo labels are generated from the network that is being trained. This allows the network to self-train from the improved pseudo labels as they are learnt. 

% Additionally, while traditional self-training segmentation methods \cite{bucher2020buda} threshold on the probabilities within each image and retrain using only those pixels which have a high prediction confidence, we observe that such an approach for road segmentation worsens the problem of loss of context.

The network in this stage is initialized with the weights obtained in Step 1. The inputs to the network are images from the target domain. Pseudo labels are generated in an online fashion from the network being trained as follows,
    \[ Y_{\textrm{pseudo}} = \begin{cases}
        1 \ \textrm{if} \ P^{h,w} \geq \tau, \\
        0 \ \textrm{otherwise},
        \end{cases}
    \]
    
\noindent where $P^{h,w}$ is the probability that a pixel belongs to the class `road' at a given location and $\tau$ is a threshold. The network is optimized with these pseudo-labels using a binary cross entropy loss term, (similar to Equation \ref{eq: cross_entropy}).

%Pixels with a prediction probability higher than $\tau$ are classified as road (class $1$), and vice versa. $\tau$ is typically set at $0.5$. 
\begin{table}
% \footnotesize
\centering
% \begin{center}
% \resizebox{.8\columnwidth}{!}{
\resizebox{\columnwidth}{!}{
\begin{tabular}{r c c}
\toprule
Dataset & Syn./Real & Weather \\
\midrule
Rainy CityScapes \cite{halder2019physics}& Syn & Varying intensities (1mm - 200mm) of rain \\
Foggy CityScapes \cite{halder2019physics}& Syn & Varying intensities (750m - 30m) of fog \\
Foggy Zurich \cite{sakaridis2018model}& Real & Light and medium Fog \\
Dark Zurich \cite{sakaridis2019guided}& Real & Twilight, Night \\
Raincouver \cite{tung2017raincouver}& Real & Rain, night \\
BDD \cite{yu2018bdd100k}& Real & Snow, Fog, Rain, Night \\

\bottomrule
\end{tabular}

} %For resize box
\caption{\textbf{List of datasets:} The second column categorizes the datasets as synthetic or real and the third column describes the images contained in the dataset.}
\vspace{-15pt}

\label{tab:datasets}
% \end{center}
\end{table}

\subsection{Few-Image Fine-Tuning via Model Distillation }
\label{sec:fewimage}
Some heterogeneous weather datasets like Raincouver \cite{tung2017raincouver} and Berkeley Deep Drive (BDD) \cite{yu2018bdd100k} contain a mixture of adversities within the same image (for instance night+rain in Raincouver, see Table \ref{tab:datasets}). Furthermore, these datasets are captured from different geographic conditions (i.e., source and target datasets may be from different regions). To make our model robust against such factors, we use ground truth labels for a few images (order of $5-10$ images) from the target dataset in a final refinement step described below. In a nutshell, given a model trained on the unlabeled target dataset using \model, and $k \leq 10$ labeled images from the target domain images, our goal is to learn enhanced feature maps for the target domain in the presence of adversarial factors such as mixtures of adversities and different geographical regions.  
% This steep variation within the datasets can make it extremely difficult for the model to learn robust representations, in addition to the domain shift caused by the change in geographical conditions (source and target datasets may be from different countries).
% The availability of ground truth labels for a few images (order of $5-10$ images) from the target dataset has the potential to greatly boost performance. 
%For instance, BDD contains images captured under snow, rain, fog, street light, glare, etc. Similarly, Raincouver contains images caputred under rain and during the night. This steep variation within datasets can make it extremely difficult for the model to capture domain specific characteristics, 
% Thus, given a model $M_{unsup}$ trained on the unlabeled target dataset using \model~, and $k$ ($k \leq 10$) labeled images from the target domain images, the goal is learn enhanced feature maps for the target domain.  

We empirically observe that directly fine-tuning the \model~model on the $k$ images is sub-optimal due to overfitting. To prevent overfitting, we propose a model distillation~\cite{li2020model} regularizer. % We show how regularization by model distillation \cite{li2020model} in concurrence with training by fine-tuning can greatly
Let the weights of the \model~model be denoted by $\omega_{\textrm{\model}}$, and the weights of the model being currently trained be denoted by $\omega_{\textrm{fewIm}}$. The cost function for model distillation is given by,
\[
    L_{\textrm{model-distil}} = C(\omega_{\textrm{\model}},\omega_\textrm{fewIm}),
\]
\noindent where $C$ represents a distance function such as MSE distance or L1 distance. In our benchmarks,  MSE distance works best. 

The network is first initialized with weights of the \model~model. The model distillation term $L_{\textrm{model-distil}}$ with weight parameter $\lambda_{\textrm{model-distil}}$ is applied in conjunction with the binary cross entropy loss function (Equation \ref{eq: cross_entropy}) to constrain the probability predictions and ground-truth labels for $k$ images. The $\lambda_{\textrm{model-distil}}$ term balances between extracting domain specific characteristics from the $k$ images (such as mix of adverse weather, geographical features etc.) and prevents the weights of the model from diverging from the \model~weights (for better generalization). The overall equation follows as,

\begin{equation}
    \mathcal{L}_{\textrm{overall}} = L_{\textrm{CE}} + \lambda_{\textrm{model-distil}} L_{\textrm{model-distil}}
\end{equation}
%
% which already contain sufficient knowledge about the domain thereby improving generalization. CAN YOU PRESENT THESE CONSTRAINTS USING EQUATIONS. Thus, a careful balance between the two terms (determined by $\lambda_{\textrm{model-distil}}$) improves performance. 

\section{Experiments and Results}

%We will make all code publicly available and provide the technical implementation details in the supplementary material. YOU SHOULD MENTION A FEW SENTENCES ON HOW RESULTS WERE IMPLEMENTED 

%We describe the datasets, and the evaluation protocol in Section\ref{sec:exp_datasets}. We present our analysis of SAfE model in Section \ref{sec:analysis_safe}. We analyse the effectiveness of \model~ on synthetic, real and heterogeneous real datasets in Sections \ref{sec:exp_syn}, \ref{sec:exp_real} and \ref{sec:exp_complexreal} respectively. Further, we demonstrate the advantages of FewIm-Ft on heterogeneous datasets in Section \ref{sec:exp_complexreal}, and finally comparisons against prior work in Section \ref{sec:sota}.

%\subsection{Datasets and Evaluation Protocol}
%\label{sec:exp_datasets}
We use the CityScapes dataset as the clear weather source domain. We conduct evaluation experiments on $6$ datasets captured in adverse environmental conditions, described in Table \ref{tab:datasets}. We evaluate our model using four metrics: mean Intersection over Union (mIoU), Recall (or accuracy), Precision, and F1 score. All our models are trained using one NVIDIA GeForce GPU, and we implement the model using the PyTorch framework. We will make all code publicly available. The hyperparameters generalize across our experiments on all datasets. For the segmenatation model, we use the SGD optimizer with a learning rate of $2.5e-4$, and momentum of $0.9$ and weight decay of $0.0005$. Dataset specific details are provided in the table below. Images are downsampled (by a factor of $2$, where necessary) by bilinear sampling, and the corresponding ground-truth labels are downsampled by nearest neighbour downsampling.

In this section, we highlight our main results which we summarize as follows,

\begin{itemize}[nosep]
    \item The self-attention auto-encoder is comparable ($94.7\%-101.25\%$ of second best SOTA mIoU) to more complex and sophisticated architectures for road segmentation (Tables~\ref{tab:cityscapesclearweather_modelcomplexity} and~\ref{tab:sup_weatherdatasets}). 
    \item We empirically show that our method approximates supervised learning-based models ($88-96\%$ of supervised mIoU) across all $6$ datasets (Tables \ref{tab:syntheticrain_mainresults},\ref{tab:syntheticrain_mainresults},\ref{tab:Foggy_Zurich},\ref{tab:Dark_Zurich},\ref{tab:raincouver},\ref{tab:bdd}).
    \item We demonstrate an improvement of at least $10.26\%$ over prior work in SFDA (Table \ref{tab:sota}).
    \item We improve training time over prior SFDA approaches by $18-180\times$.
\end{itemize}

\subsection{Analysing Pre-Trained Self-Attention-based AutoEncoder}
\label{sec:analysis_safe}
\begin{table}
%\footnotesize
\centering
% \begin{center}
\resizebox{.8\columnwidth}{!}{
\begin{tabular}{r c c}
\toprule
Model & Acc.(\%) (w/o. SA) &  Acc.(\%) (w. SA) \\
\midrule
DeepLabv2~\cite{chen2017deeplab}& $89.59$ & $90.54$ $(+0.95)$ \\
DeepLabv2 (E-1)~\cite{chen2017deeplab} & $87.50$ & $90.78$ $(+3.28)$  \\
DeepLabv2 (E-2)~\cite{chen2017deeplab} & $88.13$ & $88.62$ $(+0.49)$\\
DRN-D-105~\cite{yu2017dilated}& $83.92$ & $85.32$ $(+1.40)$\\
\textbf{DRN-D-38}~\cite{yu2017dilated}& $\textbf{90.69}$ & $\textbf{91.44}$ $(+0.75)$\\
%DRN-D-38 & CA & $88.93$ & $94.05$ & $94.22$ & $94.14$ \\
%DRN-D-54 & & $89.3$ & $94.6$ & $94.16$ & $94.38$ \\
%DRN-D-54 & BoT & $76.55$ & $85.82$ & $87.63$ & $86.72$ \\
\bottomrule
\end{tabular}
}
\caption{\textbf{Effect of self-attention:} We show for various backbone architectures that self-attention improves the accuracy. DeepLabv2 (E-1) and DeepLabv2 (E-2) denote the removal of $1$ and $2$ layers from DeepLab respectively. We select DRN-D-38 with self-attention (\textbf{bolded}) as the baseline for all further experiments.}
\vspace{-12pt}

\label{tab:cityscapesclearweather_modelcomplexity}
% \end{center}
\end{table}

\begin{comment}
\begin{table}
%\footnotesize
\centering
% \begin{center}
\resizebox{.8\columnwidth}{!}{
\begin{tabular}{c c c c c c}
\toprule
Model & Attn. & mIoU & Recall & Prec. & F1 \\
\midrule

DeepLabv2 & & $89.59$ & $96.11$ & $92.97$ & $94.51$\\
DeepLabv2& \checkmark & $90.54$&	$96.11$&	$93.98$&	$95.04$\\
\midrule
DeepLabv2 (E-1) & & $87.50$ & $94.08$ & $92.6$ & $93.33$\\
DeepLabv2 (E-1) & \checkmark & $90.78$ & $96.05$ & $94.3$ & $95.17$\\
\midrule
DeepLabv2 (E-2) & & $88.13$ & $94.19$ & $93.22$ & $93.69$\\
DeepLabv2 (E-2) & \checkmark & $88.62$ & $94.54$ & $93.39$ & $93.96$\\
\midrule
DRN-D-105 & & $83.92$ & $90.33$ & $92.2$ & $91.26$\\
DRN-D-105 & \checkmark & $85.32$ & $92.14$ & $92.01$ & $92.07$\\
\midrule 
DRN-D-38 & & $90.69$ & $95.46$ & $94.77$ & $95.11$ \\
\textbf{DRN-D-38} & \checkmark & \textbf{91.44} & \textbf{96.34} & \textbf{94.72} & \textbf{95.52}\\
%DRN-D-38 & CA & $88.93$ & $94.05$ & $94.22$ & $94.14$ \\
%DRN-D-54 & & $89.3$ & $94.6$ & $94.16$ & $94.38$ \\
%DRN-D-54 & BoT & $76.55$ & $85.82$ & $87.63$ & $86.72$ \\
\bottomrule
\end{tabular}
}
\caption{\textbf{Effect of self-attention:} We show for various backbone architectures that self-attention improves the accuracy. We select DRN-D-38 with self-attention (\textbf{bolded}) as the baseline for all further experiments.}
\vspace{-12pt}

\label{tab:cityscapesclearweather_modelcomplexity}
% \end{center}
\end{table}
\end{comment}

%\input{Tables/Supervised/ablations_clearcityscapes}
\begin{table}
% \footnotesize
\centering
% \begin{center}
%\resizebox{.8\columnwidth}{!}{
\resizebox{.8\columnwidth}{!}{
\begin{tabular}{c c c c c}
\toprule
Intensity & mIoU & Recall & Prec. & F1 \\
\midrule
\multicolumn{5}{c}{I. Synthetic Rain \cite{halder2019physics}}\\
\midrule
$1$mm & $93.23$	& $96.13$ &	$96.86$ &	$96.49$\\
$5$mm & $94.06$ &	$96.95$ &	$96.92$ &	$96.93$\\
$17$mm & $93.92$&	$96.87$&	$96.85$&	$96.86$\\
$25$mm & $93.07$&	$96.4$&	$96.42$&	$96.41$\\
$50$mm & $92.45$&	$95.72$&	$96.43$&	$96.08$\\
$75$mm & $92.05$ &	$96.25$ &	$95.47$ &	$95.86$\\
$100$mm & $91.57$ &	$96.66$ &	$94.56$ &	$95.60$\\
$200$mm & $90.57$ &	$95.69$ &	$94.42$ &	$95.05$ \\
\midrule
\multicolumn{5}{c}{II. Synthetic Fog \cite{halder2019physics}}\\
\midrule
$750$m & $95.53$&	$97.46$&	$97.97$&	$97.71$\\
$375$m & $94.74$	& $97.08$&	$97.51$&	$97.30$\\
$150$m & $92.72$&	$95.71$&	$96.74$&	$96.22$\\
$75$m & $91.64$&	$96.12$&	$95.15$&	$95.63$\\
$50$m & $90.59$&	$94.97$ &	$95.16$&	$95.06$\\
$40$m & $90.21$ &	$95.68$	& $94.03$	&$94.85$\\
$30$m & $89.00$	& $94.33$	& $94.02$	& $94.18$\\
\midrule
\multicolumn{5}{c}{III. Real datasets}\\
\midrule 
Raincouver \cite{tung2017raincouver} & $71.88$&	$80.36$&	$87.19$&	$83.64$\\
BDD \cite{yu2018bdd100k}& $89.19$&	$92.86$&	$95.76$&	$94.29$\\
% \midrule
\midrule 
\multicolumn{5}{c}{IV. SOTA Comparisons}\\ 
\midrule 
Dataset & Method & mIoU & Recall & F1 \\
\midrule 
 \multirow{8}{*}{CityScapes}
 & FCN \cite{long2015fully} & $89.90$ & $95.70$ &  $94.68$\\
 & CA \cite{fu2019dual}& $88.93$ & $94.05$ &  $94.14$ \\
 & BoT \cite{srinivas2021bottleneck} & $76.55$ & $85.82$ &  $86.72$ \\
 & DeepLabv3 \cite{chen2017rethinking}&  $90.87$ & $95.35$ &  $95.21$\\
 & s-FCN-loc \cite{wang2017embedding} &  $91.04$ & $96.11$ &  $95.36$\\
 & Zohourian \cite{zohourian2018superpixel} & $86.34$ & \textcolor{blue}{96.76} &  $92.44$ \\
 & \textbf{RBA} \cite{sun2019reverse}  & \textbf{96.00} & \textbf{98.13} &  \textbf{98.00}\\
 & Ours & \textcolor{blue}{91.44} & 96.34 & \textcolor{blue}{95.52}\\
\midrule 
\multirow{2}{*}{100mm Rain}
 & SNE-Seg \cite{fan2020sne}& \textcolor{blue}{90.80} & \textbf{96.80} &  \textbf{95.80}\\
 & Ours & \textbf{91.57} &	\textcolor{blue}{96.66} &	\textcolor{blue}{95.60}\\
\midrule
\multirow{2}{*}{100mm Rain}
 & SNE-Seg \cite{fan2020sne}& \textcolor{blue}{90.50} & \textcolor{blue}{93.00} &  \textcolor{blue}{94.47}\\
 & Ours & \textbf{91.64} &	\textbf{96.12} &		\textbf{95.63}\\

\bottomrule
\end{tabular}
 }
\caption{\textbf{Benchmarking and comparing the self-attention based pre-trained model:} We train our self-attention model from Section~\ref{sec:sup} on various datasets in a supervised manner. These supervised numbers help us conduct a relative study of the performance of \model~, which is unsupervised in the target domain. Additionally, we observe that our pre-trained model is comparable to the state-of-the-art (Experiment IV). Best results are in \textbf{bold} fonts, second best results are in \textcolor{blue}{blue}). 
}
\vspace{-12pt}

\label{tab:sup_weatherdatasets}
% \end{center}
\end{table}

In this section, we analyse the performance of our self-attention auto-encoder model (described in Section \ref{sec:sup} and benchmark its performance on various datasets in the supervised setting. In Table \ref{tab:cityscapesclearweather_modelcomplexity} (I), we show that the usage of self-attention within various conventional semantic segmentation models to encode semantic relationships via capturing long-range dependencies improves performance over the corresponding baseline. We use the DRN-D-38 model \cite{yu2017dilated} with self-attention in all further experiments. In Table \ref{tab:sup_weatherdatasets} (I,II,III), we benchmark the model on various weather datasets in the supervised setting. We use these supervised numbers in the following subsections to perform a comparative study of our self-supervised model (See Figure \ref{fig: coverpic} for a summary). In Table \ref{tab:sup_weatherdatasets} (IV), we show that the self-attention autoencoder is comparable to the state-of-the-art on various datasets.

\begin{table}
% \footnotesize
\centering
% \begin{center}
 \resizebox{.8\columnwidth}{!}{
\begin{tabular}{c c c c c c}
\toprule
Intensity & Experiment & mIoU & Recall & Prec. & F1 \\
\midrule
\multicolumn{6}{c}{I. Synthetic Rain}\\
\midrule
1 mm & A & 94.73&	97.16&	97.42&	97.29\\
5 mm & A & 94.09&	97.01&	96.90&	96.95\\
17 mm & A & 93.95&	96.83&	96.93&	96.88\\
25 mm & A & 93.01&	96.30&	96.46&	96.38\\
50mm & A & 89.82 &	93.71&	95.58&	94.64\\
\midrule
%\multicolumn{6}{c}{II. 75 mm Rain}\\
75mm & A & 87.15 &	90.92 & 	95.45 &	93.13\\
75mm & B & 88.08 &	93.94 &	93.38 &	93.66 \\
\midrule
%\multicolumn{6}{c}{III. 100 mm Rain}\\
100mm & A & 82.97&	86.88&	94.84&	90.69\\
100mm & B& 86.67&	92.88&	92.84&	92.86 \\
\midrule
%\multicolumn{6}{c}{IV. 200 mm Rain}\\
200mm & A & 65.98 &	68.94 &	93.90 &	79.50\\
200mm & B & 80.25 &	86.96 &	91.22 &	89.04 \\
%C, from 75mm rain & 81.09 &	88.15 &	91.00 &	89.56 \\
%C, from 100mm rain & 81.58 &	88.55 &	91.2 &	89.85\\
200mm & C & 81.58 &	88.55 &	91.20 &	89.85\\
\midrule
\multicolumn{6}{c}{II. Synthetic Fog}\\
\midrule
750m & A & 94.90	& 96.68&	98.09&	97.38\\
375m & A & 92.71 &	94.60&	97.88&	96.21\\
\midrule
%\midrule
%\multicolumn{5}{c}{Fog 150m}\\
150m & A & 85.55&	87.57&	97.38&	92.21\\
150m & B & 89.74&	94.59&	94.59&	94.59\\
%\midrule
\midrule
%\multicolumn{5}{c}{Fog 75m}\\
75m & A & 70.81 &	72.51 &	96.79 &	82.9\\
75m & B & 80.67 &	83.01 &	96.62 &	89.30\\
75m & C & 87.48&	92.99&	93.65&	93.32\\
%\midrule
\midrule
%\multicolumn{5}{c}{Fog 50m}\\
50m & A & 57.29 &	58.92&	95.39&	72.85\\
50m & B & 71.64&	73.69&	96.24&	83.47\\
50m & C & 85.47&	90.75&	93.63&	92.16\\
\midrule
%\midrule
%\multicolumn{5}{c}{Fog 40m}\\
40m & A & 46.27&	48.45&	91.12&	63.27\\
40m & B & 52.62 &	53.94&	95.57&	68.96\\
40m & C & 83.65 &	89.19&	93.09&	91.10\\
%\midrule
\midrule
%\multicolumn{5}{c}{Fog 30m}\\
30m & A & 35.04&	38.05&	81.59&	51.90\\
30m & B & 22.24&	22.92&	88.29&	36.39\\
30m & C & 80.82 &	85.96&	93.11&	89.39\\

\bottomrule
\end{tabular}
 }
\caption{\textbf{Results on synthetic rain and synthetic fog \cite{halder2019physics}:} We evaluate our model under three settings - A: (Baseline) Testing the clear weather CityScapes model, B: \model~without curriculum learning, C: \model. We show that the CityScapes model generalizes well to light synthetic rain and fog, and its performance degrades as the intensity of rain and fog increases, which is restored by \model. 
\textbf{Comparison to supervised results:} On synthetic rain, \model~achieves 90.07\%-90.8\% and 92.53\%-97.6\% of supervised mIoU and recall, respectively. On synthetic fog, \model~ achieves 90.8\%-96.78\% and 91.12\%-98.82\% of supervised mIoU and recall respectively. }
\label{tab:syntheticrain_mainresults}
\vspace{-12pt}

% \end{center}
\end{table}

\subsection{Results on Synthetic Datasets: Rain and Fog}
\label{sec:exp_syn}
We perform three evaluation experiments. Experiment A corresponds to testing the pre-trained CityScapes model on varying intensities of rain and fog. Experiment B corresponds to results obtained by training \model~without curriculum learning (\textit{i.e.} by initializing with the CityScapes pre-trained model and training on higher intensities of rain and fog directly), and experiment C corresponds to results obtained by training \model~as proposed in Section \ref{sec:unsup} (\ie with curriculum learning).
%\paragraph{Results on Synthetic Rain:}

\noindent \textbf{Results on Synthetic Rain:} The results of \model~ on Synthetic Rainy CityScapes \cite{halder2019physics} are shown in Table \ref{tab:syntheticrain_mainresults} (I). For low intensities of rain, the performance of the CityScapes model is more or less preserved (Table~\ref{tab:sup_weatherdatasets} (I)). For higher intensities of rain, there is a degradation in performance. We observe that the decrease in performance is highest for $200$mm rain, at $27.15\%$.  For $75$mm, $100$mm, and $200$mm, Experiment B imparts an improvement of $1.03\%$, $4.45\%$, and $21.63\%$ respectively over the corresponding baselines. Experiment C leads to a cumulative improvement of $23.64\%$ over the corresponding baseline. Furthermore, we demonstrate that the mIoU and recall of our \model, which is completely unsupervised, is $90.07\%-90.80\%$ and $92.53\%-97.60\%$ of the counterpart supervised mIoU. 

%\paragraph{Results in Synthetic Fog:} 
\noindent \textbf{Results on Synthetic Fog:} The results of our \model~ on Synthetic Foggy CityScapes \cite{halder2019physics} are shown in Table \ref{tab:syntheticrain_mainresults}II. Similar to synthetic rain, the performance of the CityScapes model on light fog is more or less preserved (Table \ref{tab:sup_weatherdatasets} (II)). For higher intensities of fog, there is a degradation in performance. We notice that the degradation is very high for visibility distances less than 150m, and is the highest at $60.6\%$ for 30m fog. Experiment B leads to an improvement of 4.89\%, 13.9\%, 25.04\%, 13.7\% on 150m, 75m, 50m, and 40m fog respectively. Direct application of the self-training algorithm on 30m fog (without curriculum learning) degrades performance since the generalization of CityScapes on 30m is very poor. Experiment C (Training SelfTr-Road Seg by curriculum learning i.e. progressively from 150m fog to 30m fog) cumulatively improves performance over the baselines (Experiment A) by 23.54\%, 28.18\%, 80.78\% and 130.65\% on 75m, 50m 40m, and 30m fog respectively. Furthermore, we demonstrate that the mIoU and recall of our \model, which is completely unsupervised, is 90.8\%-96.78\% and 91.12\%-98.82\% of the supervised mIoU. 

%\paragraph{The effectiveness of curriculum learning:} 
\noindent \textbf{Benefits of Curriculum Learning:} On a given dataset, the performance of our self-training algorithm depends heavily on the generalization capabilities of the pre-trained model used for initialization. Therefore, progressively initializing and training the model on increasing intensities of rain and fog will lead to the best accuracies (Table \ref{tab:syntheticrain_mainresults}, Table \ref{tab:syntheticrain_mainresults}). This implies that progressively training from low intensities to high intensities of rain and fog improves the quality of pseudo labels and prediction probabilities on high intensities of rain and fog. In simpler terms, for the $100$mm rain dataset, a model trained on $75$mm rain will work better than the CityScapes clear weather model. Similarly, for the $200$mm rain dataset, a model trained on $100$mm rain will work better than the $75$mm rain model, which in turn will generalize better the CityScapes clear weather model. A similar intuition can be drawn for synthetic fog too. We validate this hypothesis in Table \ref{tab:clpseudolabels_proof}. The first and second columns correspond to the datasets that the model is trained and tested on respectively. We observe that curriculum learning progressively improves performance, which results in high quality pseudo labels with high confidence, a boon for self-training.

%Consider evaluating the performance on 75m fog. A model trained on 150m fog works better than the CityScapes clear weather fog. Similarly, when evaluating on 50m fog, we demonstrate that a model trained on 75m fog works better than a model trained on 150m fog, which in turn performs better than the CityScapes clear weather model. The trends scale similarly for higher intensities of fog too. Thus, self-training by initialization with these models with enriched pseudo labels leads to enhanced performance.

\noindent \textbf{Generalization trends:} We observe that our self-supervised \model~preserves the accuracy on the clear weather source dataset, CityScapes. The mIoU and accuracy of the fog model on clear weather CityScapes are at $96.9\%$ and $99.00\%$ of the supervised CityScapes model, respectively. The corresponding accuracy numbers for synthetic rain are  $98.11\%$ and $98.72\%$, respectively. %(IoU-Rain : 89.72	R-95.11	P-94.05	F1-94.58; (IoU-88.65	R -95.38	P-92.64	F-93.98)). 

\begin{table}
% \footnotesize
\centering
% \begin{center}
\resizebox{.8\columnwidth}{!}{
\begin{tabular}{c c c c c c}
\toprule
Training & Testing & mIoU & Recall & Prec. & F1 \\
\midrule
%\multicolumn{5}{c}{I. Tested on 100 mm rain}\\
CS & 100mm rain & 82.97&	86.88&	94.84&	90.69\\
75mm rain & 100mm rain & 86.31 &	92.63 &	92.68 &	92.65 \\
\midrule
%\multicolumn{5}{c}{II. Tested on 200 mm rain}\\
CS & 200mm rain & 65.98 &	68.94 &	93.9 &	79.5\\
75 mm rain& 200mm rain & 78.97 &	85.86 &	90.77 &	88.25\\
100 mm & 200mm rain & 79.61	& 86.27	& 91.15	& 88.65\\
\midrule
%\multicolumn{5}{c}{III. Tested on 75m fog}\\
CS & 75m fog & 70.81 &	72.51 &	96.79 &	82.9\\
150m fog & 75m fog & 83.82 &	87.44 &	95.28&	91.19 \\
\midrule 
%\multicolumn{5}{c}{IV. Tested on 50m fog}\\
CS & 50m fog & 57.29 &	58.92&	95.39&	72.85\\
150m fog & 50m fog & 76.07&	79.49&	94.64&	86.41\\
75m fog & 50m fog & 84.38	& 89.44	& 93.71	& 91.52\\
\midrule
%\multicolumn{5}{c}{V. Tested on 40m fog}\\
CS & 40m fog & 46.27&	48.45&	91.12&	63.27\\
150m fog & 40m fog & 65.6&	69.98&	91.28&	79.22\\
75m fog & 40m fog & 78.17	& 83.51	& 92.43	& 87.74\\
\midrule
%\multicolumn{5}{c}{VI. Tested on 30m fog}\\
CS & 30m fog & 35.04&	38.05&	81.59&	51.9\\
150m fog & 30m fog & 53.61	& 59.62	& 84.17	& 69.8\\
75m fog & 30m fog & 68.49	& 75.21	& 88.45	& 81.3\\
\bottomrule
\end{tabular}
}

\caption{\textbf{Curriculum learning improves performance by improving the accuracy of pseudo labels:} The first and second columns correspond to the datasets our model is trained and tested, respectively. Evaluating on 30m fog reveals that a model finetuned to 75m fog performs better than the counterpart 150m fog model, which in turn performs better than that the CityScapes model. Similar conclusions can be drawn for other intensities of rain/fog.}
\label{tab:clpseudolabels_proof}
\vspace{-12pt}

% \end{center}
\end{table}

%Dark Zurich - 2
%Foggy Zurich - 4
%Rain100mm - 3
%Fog 50m - 1

\begin{figure*}[t]
    \centering
    % \captionsetup[subfigure]{labelformat=empty}
    \begin{subfigure}[b]{0.12\textwidth}
    \includegraphics[scale=0.08]{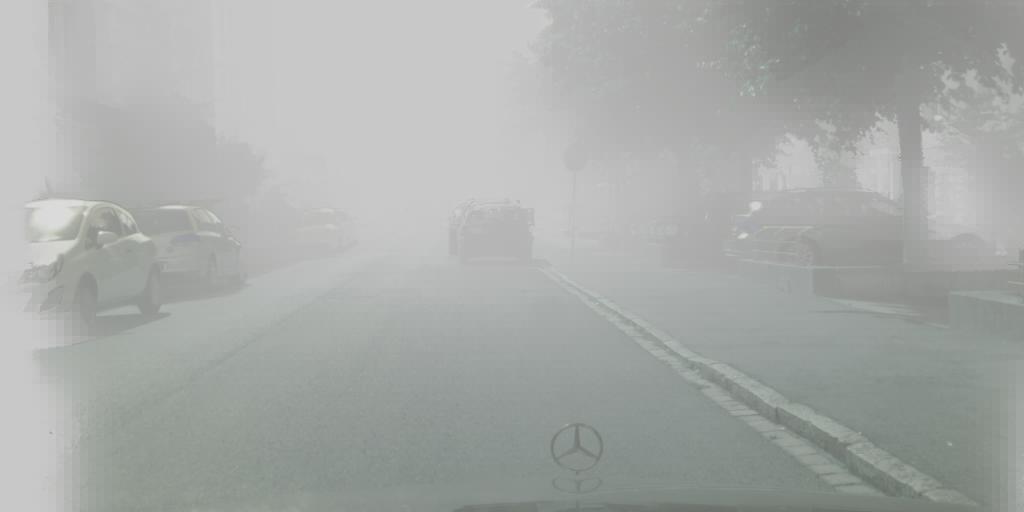}
    \caption{Fog 50m}
    \end{subfigure}
    \begin{subfigure}[b]{0.12\textwidth}
    \includegraphics[scale=0.08]{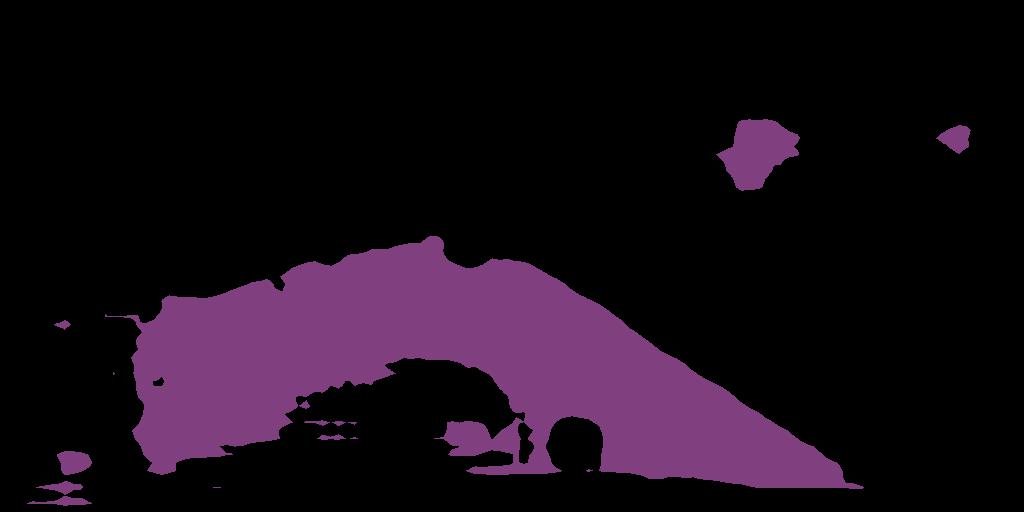}
    \caption{Baseline}
    \end{subfigure}
    \begin{subfigure}[b]{0.12\textwidth}
    \includegraphics[scale=0.08]{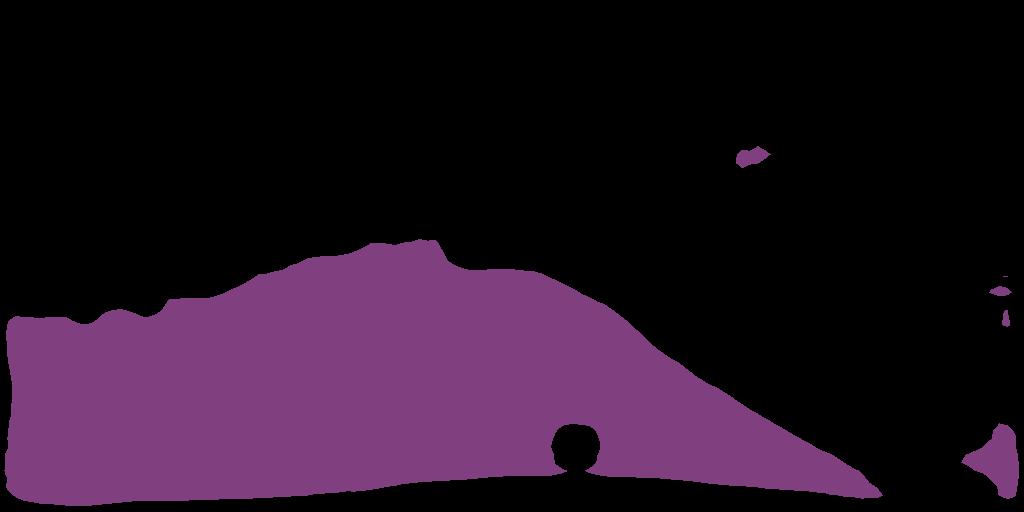}
    \caption{Ours}
    \end{subfigure}
    \begin{subfigure}[b]{0.12\textwidth}
    \includegraphics[scale=0.08]{{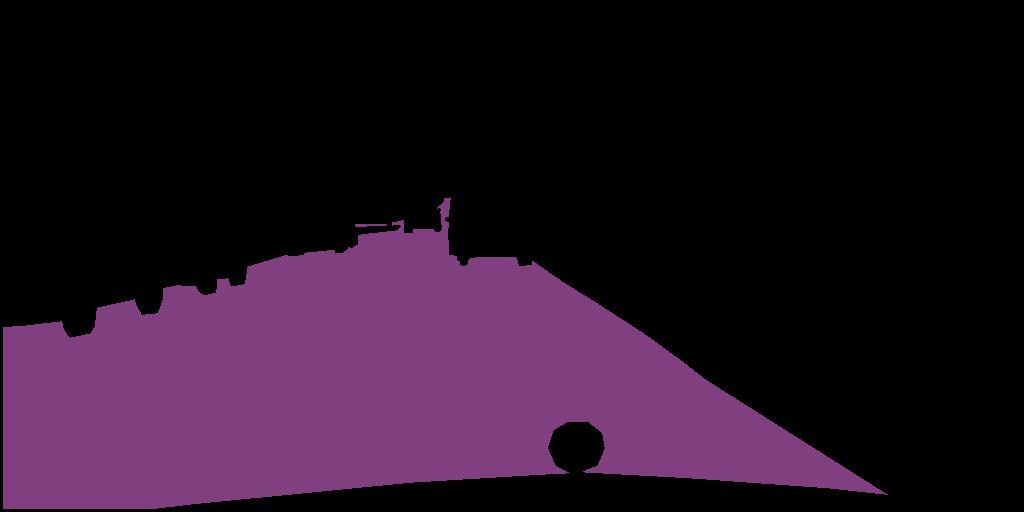}}
    \caption{GT}
    \end{subfigure}
    \begin{subfigure}[b]{0.12\textwidth}
    \includegraphics[scale=0.08]{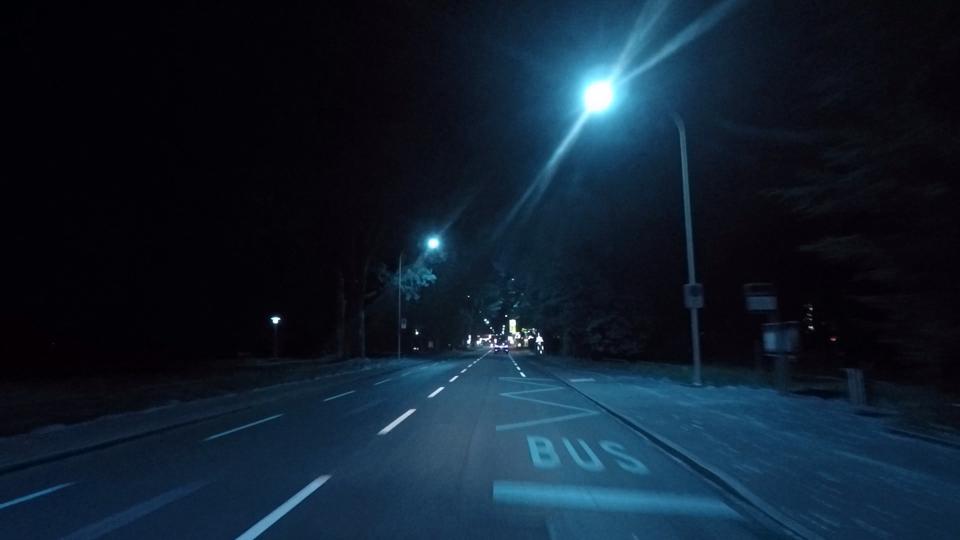}
    \caption{Dark Zurich}
    \end{subfigure}
    \begin{subfigure}[b]{0.12\textwidth}
    \includegraphics[scale=0.08]{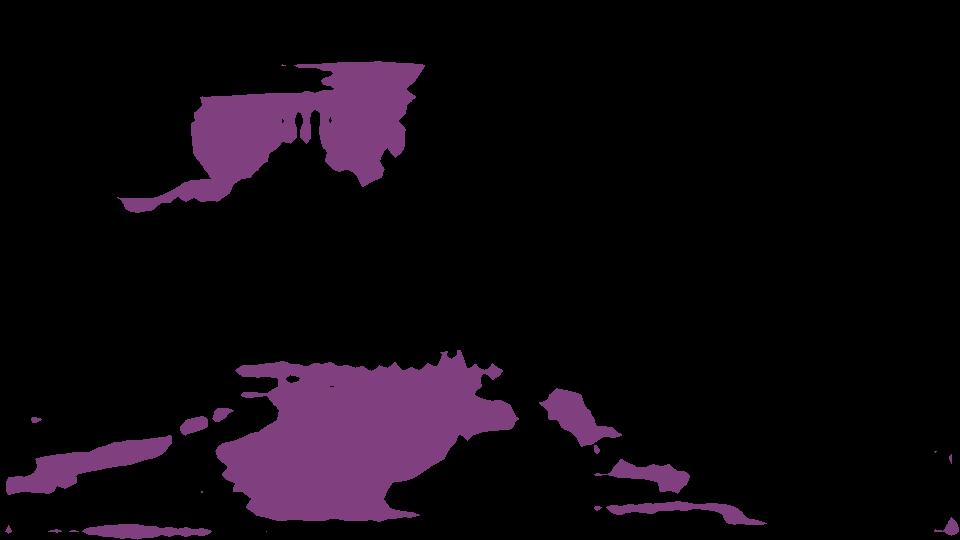}
    \caption{Baseline}
    \end{subfigure}
    \begin{subfigure}[b]{0.12\textwidth}
    \includegraphics[scale=0.08]{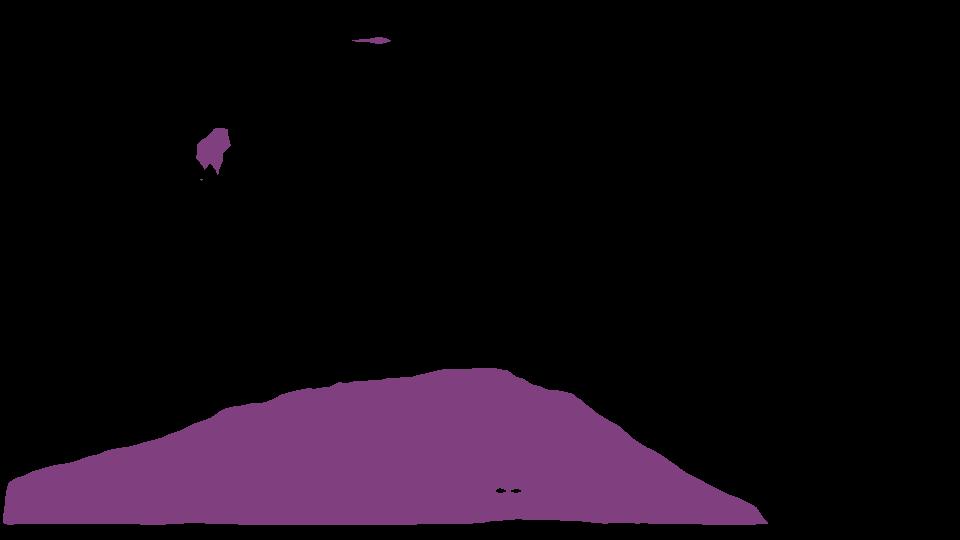}
    \caption{Ours}
    \end{subfigure}
    \begin{subfigure}[b]{0.12\textwidth}
    \includegraphics[scale=0.08]{{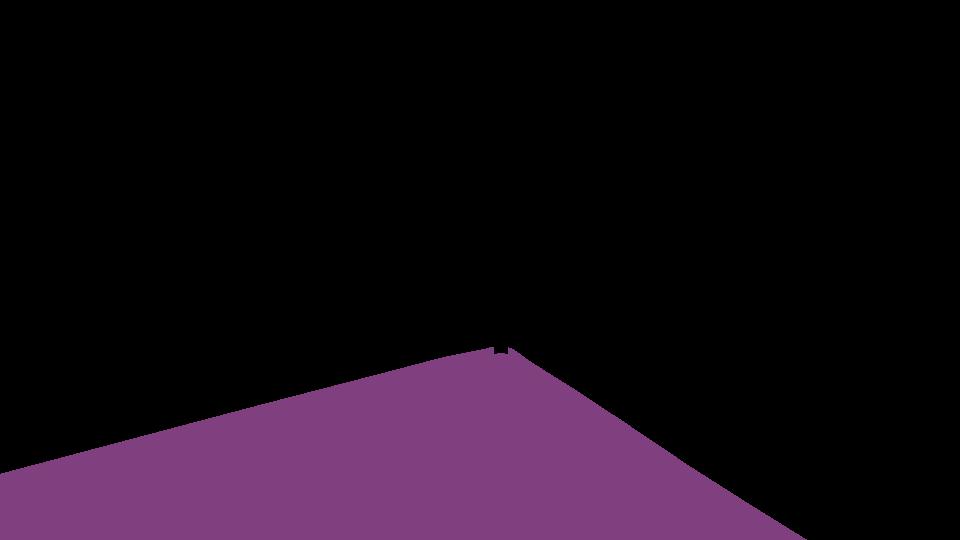}}
    \caption{GT}
    \end{subfigure}
    \\
    \begin{subfigure}[b]{0.12\textwidth}
    \includegraphics[scale=0.08]{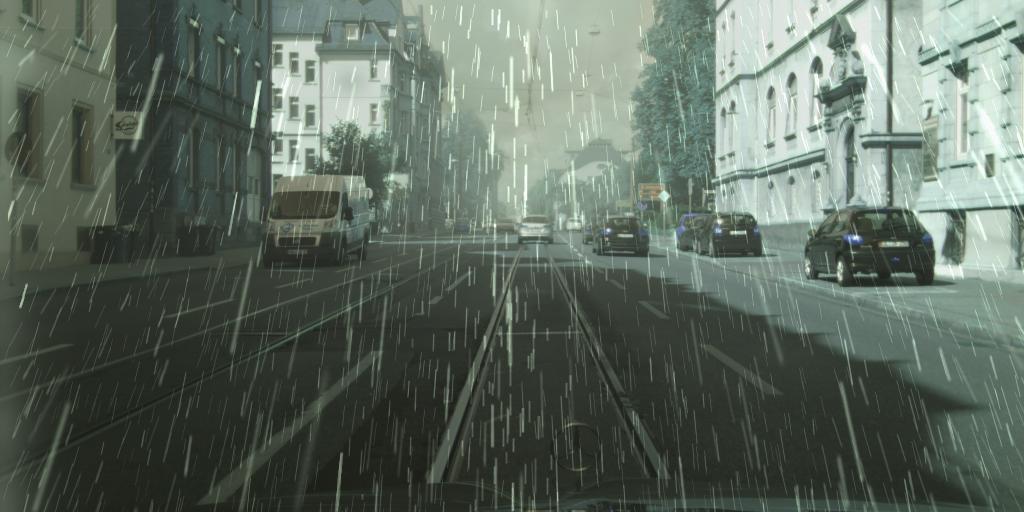}
    \caption{Rain 100mm}
    \end{subfigure}
    \begin{subfigure}[b]{0.12\textwidth}
    \includegraphics[scale=0.08]{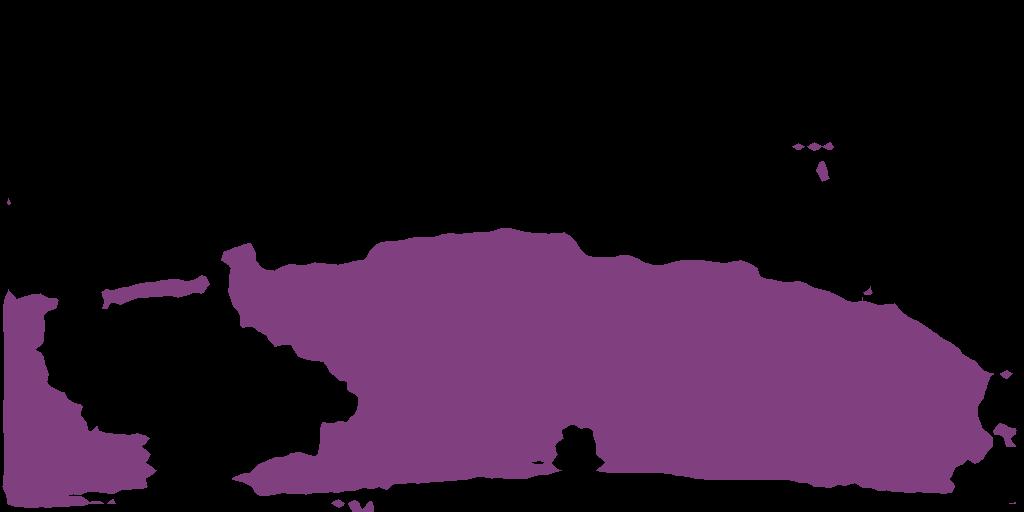}
    \caption{Baseline}
    \end{subfigure}
    \begin{subfigure}[b]{0.12\textwidth}
    \includegraphics[scale=0.08]{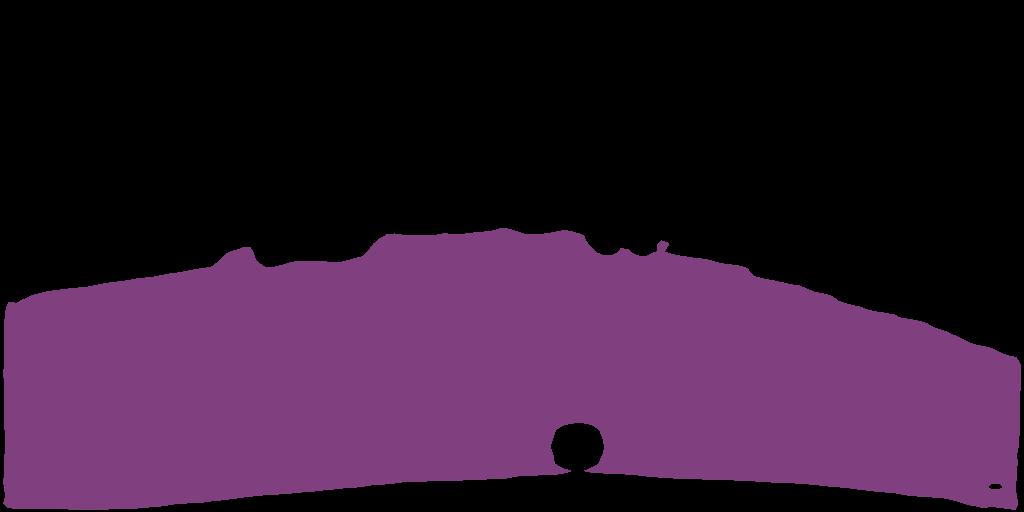}
    \caption{Ours}
    \end{subfigure}
    \begin{subfigure}[b]{0.12\textwidth}
    \includegraphics[scale=0.08]{{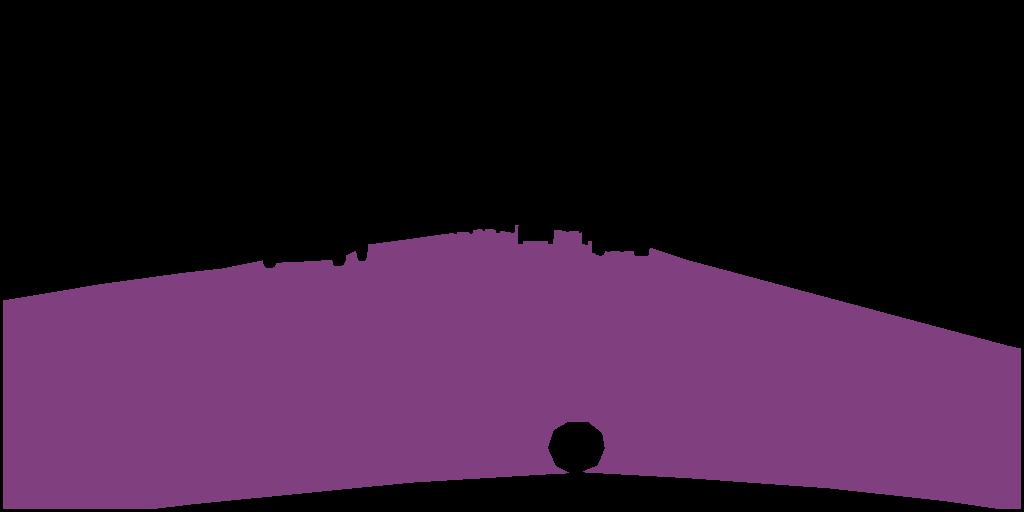}}
    \caption{GT}
    \end{subfigure}
    \begin{subfigure}[b]{0.12\textwidth}
    \includegraphics[scale=0.08]{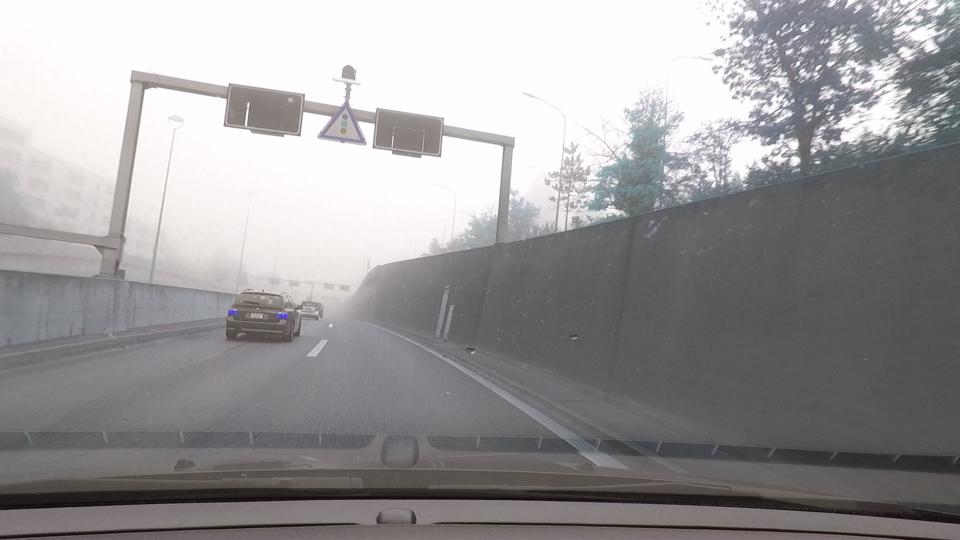}
    \caption{Foggy Zurich}
    \end{subfigure}
    \begin{subfigure}[b]{0.12\textwidth}
    \includegraphics[scale=0.08]{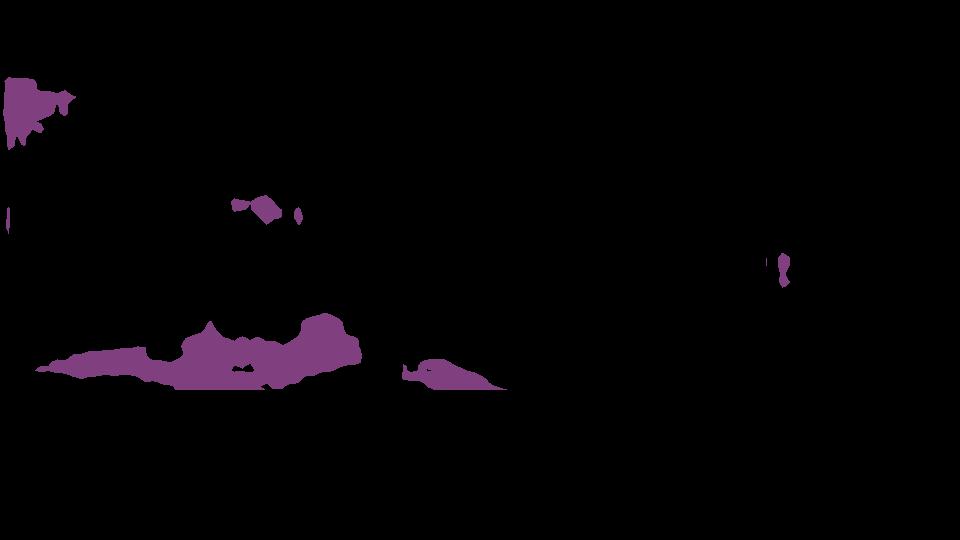}
    \caption{Baseline}
    \end{subfigure}
    \begin{subfigure}[b]{0.12\textwidth}
    \includegraphics[scale=0.08]{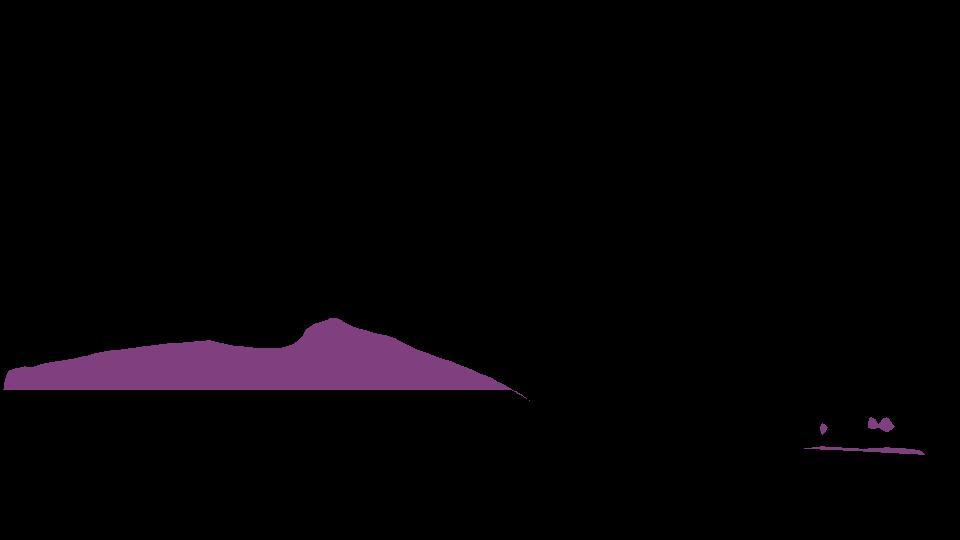}
    \caption{Ours}
    \end{subfigure}
    \begin{subfigure}[b]{0.12\textwidth}
    \includegraphics[scale=0.08]{{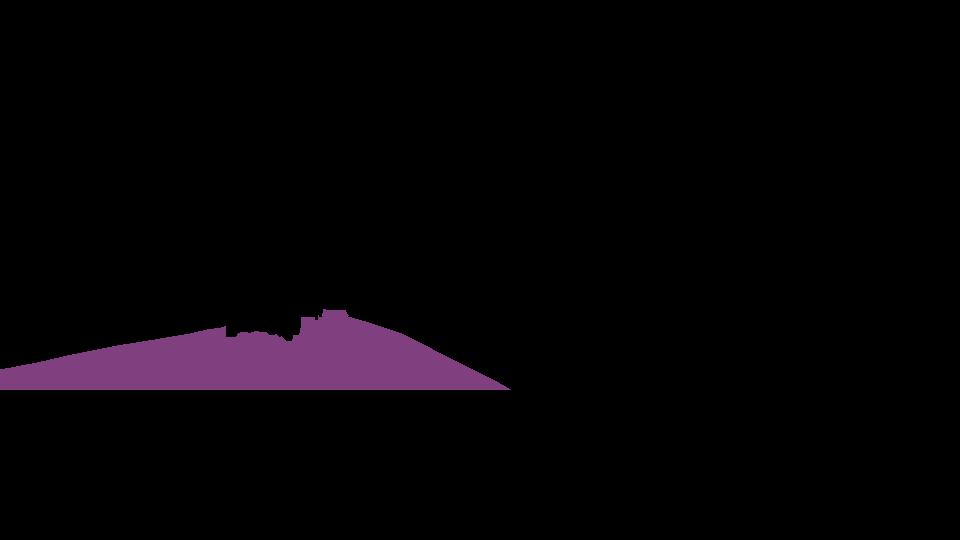}}
    \caption{GT}
    \end{subfigure}
    
    \caption{\textbf{Qualitative results.} Our model generates results that closely resemble the ground-truth (GT) compared to the baseline CityScapes pre-trained model. Purple indicates the segmented road region. More results can be found in the supplementary material.}
    \label{fig:visualisations}
    %\vspace{-10pt}
\end{figure*}

\begin{table}
%\footnotesize
\centering
% \begin{center}
\resizebox{0.8\columnwidth}{!}{
\begin{tabular}{c c c c c}
\toprule
Experiment & mIoU & Recall & Prec. & F1 \\
\midrule
\multicolumn{5}{c}{I. Baseline: CityScapes pre-trained model}\\
\midrule
- & 36.91 &	57.04 &	51.13& 53.92\\
\midrule
\multicolumn{5}{c}{II. \model~ on Light Fog; Init: CS model} \\
\midrule
Step 1 & 59.1 &	68.27 &	81.48&74.29\\
Step 2 & 60.9 &	69.09 &	83.7 &75.7\\
\midrule
\multicolumn{5}{c}{III. \model~ on Medium Fog; Init: Light fog model; }\\ \midrule
Step 1 & 73.34	& 76.22	& 95.1	& 84.62\\
Step 2 & \textbf{74.5}	& \textbf{76.74}& \textbf{96.23}& \textbf{85.39}\\

\bottomrule
\end{tabular}
}
\caption{\textbf{Results on the real fog dataset, Foggy Zurich}. We observe that \model~ improves mIoU by 101.84\% over the corresponding baseline. We notice that training the model on light fog followed by medium fog imparts the model with improvised self-training abilities, the mIoU improves by 22.3\%. Additionally, the ablations on each minibatch indicate that the two step training procedure in \model~ is indeed helpful.}
\label{tab:Foggy_Zurich}
\vspace{-12pt}

% \end{center}
\end{table}

\subsection{Results on Real datasets: Foggy Zurich and Dark Zurich}
\label{sec:exp_real}
\noindent \textbf{Analysis: Foggy Zurich} The results are presented in Table \ref{tab:Foggy_Zurich}. The pre-trained CityScapes model fails to generalize (Table \ref{tab:Foggy_Zurich} (I)) to the real fog dataset due to the domain gap. In accordance with the curriculum learning strategy, the self-training algorithm is first applied on images with light fog, and then on images with medium fog. We show results for each of the two stages of \model~to demonstrate how the two-step training procedure gradually improves performance. Training with light fog improves the mIoU by $64.99\%$ over the corresponding clear weather baseline. Initializing the model trained on light fog, and fine-tuning on medium fog using \model~further improves the performance by $22.3\%$, thus resulting in a cumulative improvement of $101.84\%$. We do not show comparisons with supervised methods due to a lack of labeled datasets.

\begin{table}
\centering
% \begin{center}
\resizebox{0.8\columnwidth}{!}{
\begin{tabular}{c c c c c}
\toprule
Experiment & mIoU & Recall & Prec. & F1 \\
\midrule
% \midrule
\multicolumn{5}{c}{I. Baseline: CityScapes pre-trained model}\\
\midrule
- & 54.9&	63.89&	79.6&	70.88\\
\midrule
\multicolumn{5}{c}{II. \model~ on Twilight Zurich; Init: CS model}\\
\midrule
Step 1 & 69.11&	74.75&	90.15&	81.74\\
Step 2 & 69.13&	74.77&	90.16&	81.74\\
\midrule
\multicolumn{5}{c}{III. \model~ on Night Zurich; Init: Twilight model}\\
\midrule
Step 1 & 71.75&	76.22&	92.44&	83.55\\
Step 2 & \textbf{72.18}&	\textbf{76.51}&	\textbf{92.72}&	\textbf{83.84}\\
% \midrule

\bottomrule
\end{tabular}
}
\caption{\textbf{Results on the night driving dataset, Dark Zurich}. We observe that \model~ improves mIoU by 31.47. We observe that our two-stage training routine within the curriculum learning algorithm helps the network perform well.}
\label{tab:Dark_Zurich}
\vspace{-12pt}

% \end{center}
\end{table}

\begin{table}
% \footnotesize
\centering
% \begin{center}
\resizebox{.7\columnwidth}{!}{
\begin{tabular}{c c c c c}
\toprule
Experiment & mIoU & Recall & Prec. & F1 \\
\midrule
I. \model~ & 52.42 &	93.96 &	54.25&	68.78\\
\midrule
\multicolumn{5}{c}{II. Results on FewIm-FT (Section \ref{sec:fewimage})}\\
\midrule 
k=1 & 54.71	& 90.7&	57.96&	70.74\\
k=2 & 56.74	& 90.45&	60.36&	72.4\\
k=10 & 63.85 &	79.39&	76.54&	77.93\\
\midrule
\multicolumn{5}{c}{III. Effect of model distillation (MD); k=10}\\
\midrule
Without MD & 57.08 &	70.99 &	74.45&	72.67\\
With MD & 63.85 &	79.39&	76.54&	77.93\\
% \midrule
\bottomrule
\end{tabular}
}
\caption{\textbf{Results on Raincouver:} \model~ achieves 72.9\% of supervised mIoU. We demonstrate the effectiveness of FewIm-FT (Section \ref{sec:fewimage}) over varying values of $k$ in Experiment II. $k=10$ results in an improvement of 21.8\% over \model~, which is 88.82\% of supervised mIoU. Additionally, we show the benefit of model distillation in Experiment III.}
\label{tab:raincouver}
\vspace{-12pt}

% \end{center}
\end{table}

\begin{table}
% \footnotesize
\centering
% \begin{center}
\resizebox{.8\columnwidth}{!}{
\begin{tabular}{c c c c c}
\toprule
Experiment & mIoU & Recall & Prec. & F1 \\
\midrule
%\multicolumn{5}{c}{I. Testing the CityScapes model on BDD}\\
%\midrule
CS pre-trained model & 70.28 &	74.49 &	92.54 &	82.54\\
%\midrule
\model~ & 75.29 &	83.3 &	88.7 &	85.92\\
%\midrule
%\multicolumn{5}{c}{III. Results on FewIm-FT}\\
%\midrule
FewIm-FT, k=10 & 83.05 &	93.06 &	88.53 &	90.74\\
%k=5 & 82.47	& 93.18	& 87.77	& 90.39\\
%k=3 & 79.3	& 89.23	& 87.68	& 88.45\\
%k=2 & 79.47	& 88.76	& 88.36	& 88.56\\
%k=1 & 71.78	& 79.27	& 88.38	& 83.57\\

\bottomrule
\end{tabular}
}
\caption{\textbf{Results on BDD:} \model~ improves baseline mIoU by 7.12\%. Experiments on FewIm-FT (Section \ref{sec:fewimage}) demonstrate an improvement of 10.30\% over \model~for k=10, which is 93.11\% of supervised mIoU.}
\label{tab:bdd}
\vspace{-12pt}

% \end{center}
\end{table}

\begin{table}
% \footnotesize
\centering
% \begin{center}
\resizebox{.9\columnwidth}{!}{
\begin{tabular}{c c c c c}
\toprule
Experiment & Method & Source Data & mIoU & \%(sup)\\
\midrule
\multicolumn{5}{c}{I. Synthetic Dataset: 100 mm rain}\\
\midrule 
Domain Adaptation & ADVENT \cite{vu2019advent} & \checkmark & 85.96 & 93.87\%\\
Domain Adaptation & PrDA \cite{kim2020domain}  & \checkmark & 86.12 & 94.04\%\\
\midrule 
Self-training & CBST \cite{zou2018unsupervised} &  & 78.64 & 85.87\%\\
%Hypothesis transfer \cite{liang2020source} &  & 86.04 \\
SFDA & PrDA \cite{kim2020domain} & & 84.77 & 92.57\%\\
SFDA & \textbf{Ours} & & \textbf{86.67} & 94.64\%\\
\midrule
\multicolumn{5}{c}{II. Heterogeneous Real Dataset: Berkeley Deep Drive}\\
\midrule
SFDA & PrDA \cite{kim2020domain} & & 75.32 & 82.43\%\\
SFDA & \textbf{Ours} & & \textbf{83.05} & 93.11\%\\

\bottomrule
\end{tabular}
}
\caption{\textbf{Comparisons against the state-of-the-art:} We adopt the state-of-the-art methods in self-training, domain adaptation, and SFDA classification to road segmentation, and present the results on Rain 100mm. We pick the best performing source-free model and conduct experiments on BDD. We observe an improvement of $10.26$\%.}
\vspace{-20pt}
\label{tab:sota}
% \end{center}
\end{table}

\noindent \textbf{Analysis-Dark Zurich:} The results are presented in Table \ref{tab:Dark_Zurich}. Table \ref{tab:Dark_Zurich} (I) shows that the CityScapes pre-trained model does not work well on Dark Zurich. \model~ first trains on the twilight images, and then on the night images. Training with twilight images improves the mIoU by $24.91\%$ over the corresponding clear weather baseline. Initializing the model with the model trained on twilight images, and fine-tuning on night images further improves performance by $4.41\%$, thus resulting in a cumulative improvement of $31.47\%$. We do not show comparisons to supervised accuracies due to the unavailability of training labels.

\subsection{Heterogeneous Real Datasets: Raincouver and Berkeley Deep Drive}
\label{sec:exp_complexreal}
Raincouver~\cite{tung2017raincouver} and Berkeley Deep Drive (BDD)~\cite{yu2018bdd100k} are complex datasets with images containing a mix of weather conditions in addition to scenes from different geographical regions. Raincouver consists of images captured under rain during the night. BDD consists of images in snow, fog, low light, glare, rain, etc. We show that \model~benefits from supervised finetuning with just $5-10$ images using the procedure discussed in Section~\ref{sec:fewimage}. The models converge in 40 iterations, which takes 2 minutes to train on one NVIDIA GeForce GPU with 11GB memory. 

\noindent \textbf{Analysis-Raincouver:} The results are shown in Table~\ref{tab:raincouver}. \model~results in an mIoU of $52.42$, which is $72.9\%$ of supervised mIoU. In Table~\ref{tab:raincouver} (II), we show the effectiveness of the fine-tuning step. As $k$ (number of supervised images used by the algorithm) increases, the performance of the model improves. Using just $1$ image ($k=1$) results in an improvement of $4.3\%$ (over \model). $k=10$ achieves $88.82\%$ of supervised mIoU. In Table~\ref{tab:raincouver} (III), we demonstrate that model distillation improves mIoU by $11.86\%$. Hyperparameter tuning on the model distillation hyperparameter $\lambda_{\textrm{model-distil}}$ reveals that a value of $1.0$ works best. %Raincouver - Model distillation parameter tuning - 0.1 gives 51.07, 1.0 gives 63.85

\noindent \textbf{Analysis-Berkeley Deep Drive:} The CityScapes model results in an mIoU of $70.28$ (baseline). Training with \model~improves performance by $7.12\%$ over the baseline, and is at $84.41\%$ of supervised IoU. In Experiment III, we pick k random images from the dataset and finetune the network using the fine-tuning step. In concurrence with our intuition, we observe that the performance improves as the number of images increases. For $k=10$, we demonstrate an mIoU improvement of $10.30\%$ over \model, which is $93.11\%$ of supervised accuracy.

\subsection{Training Time and Convergence}
The pre-training step helps our model converges in 1/6 epoch (the accuracies are similar over multiple random runs), thus bringing the training time down to 15 minutes on an NVIDIA GeForce GPU with 11GB memory. Generative SFDA models~\cite{kundu2020universal} take 30 epochs to converge, while prior work on self-training \cite{zou2018unsupervised} and SFDA~\cite{kim2020domain} applied (where feasible) to road segmentation converge in 3 epochs. Thus, we improve training time over prior SFDA approaches by $18-180 \times$.  

\subsection{Comparisons with Prior Work}
\label{sec:sota}
We adapt (where feasible) the state-of-the-art method in the following categories:  self-supervised learning \cite{zou2018unsupervised}, source-free DA\cite{kim2020domain}, and DA \cite{vu2019advent} to road segmentation for comparisons. Other methods like feature alignment by class-wise prototype learning \cite{liang2020we}, generative methods \cite{kundu2020universal,hou2020source,kurmi2021domain,yeh2021sofa,li2020model}, and models that optimize source domain class priors \cite{hacohen2019power} do not scale well to SFDA road segmentation due to the problems highlighted in Section~\ref{sec:relatedwork_da}. We performed evaluations using each of the methods on synthetic rain (Table~\ref{tab:sota} (I)) and observe that our model outperforms all prior methods. We finally train the best performing SOTA model on BDD (Table \ref{tab:sota} (II)) for comparisons on heterogeneous real datasets which carry the highest level of difficulty. On BDD our method outperforms prior methods by $10.26\%$. 

%Summary: We show that our self-attention based autoencoder baseline SAfE improves performance over several baseline architectures, and is comparable to the state-of-the-art. Experiments on 6 real and synthetic datasets reveal that our self-training algorithm achieves atleast 88.82% of supervised mIoU across datasets. In terms of accuracy, our model is comparable to supervised models. \model~ requires just one pass over the dataset. For k=10, FewIm-FT converges in 40 iterations, which takes 2 minutes to train.

\section{Conclusion, Limitations, and Future Work}
%In this paper, we propose a self-training algorithm for source free domain adaptive road segmentation in hazardous driving conditions. We establish a simple self-attention based architecture for pre-training the model on the clear weather source domain, for capturing contextual information and semantic relationships. For the unsupervised source free adaptation step, we propose a two step self-training algorithm trained by curriculum learning. To further improve performance on heterogeneous weather datasets, we propose a few-image fine-tuning algorithm that uses just 5-10 labeled mages from the target domain. We demonstrate the effectiveness of our method on 6 real and synthetic datasets via extensive experiments and ablation studies, and show that the performance of our unsupervised method is comparable to supervised models. Future work in this area can focus on extending the algorithm to semantic segmentation, object detection and other perception tasks crucial for autonomous driving. 
We propose a new method for road segmentation in adverse weather conditions using a novel self-supervised source-free domain adaptation approach. 
%Our approach proposes a two-step self-self supervised algorithm based on curriculum learning and entropy minimization. 
% To further improve performance on heterogeneous weather datasets, we propose a few-image fine-tuning algorithm that uses just 5-10 labeled mages from the target domain.
Through our evaluations on $6$ real and synthetic datasets, we show that that our self-supervised model that has access to only a pre-trained clear weather model and unlabeled target images exhibits accuracy that is comparable to completely supervised models which have access to labels for all target domain images. In addition, we exhibit benefits in terms of faster training time and state-of-the-art performance.% , \textit{(ii)} the training time is $18-180 \times$ faster than prior work, and \textit{(iii)} we outperform prior work on SFDA and self-supervised learning in terms of mIoU by atleast $10.26\%$. 

There are a few limitations of our work. Currently, our approach is designed for binary segmentation and cannot perform multi-class segmentation. Extending the current method to multi-class segmentation would generalize the approach beyond road segmentation. %Our work initiates an interesting possibility for future work in the area of SFDA. Specifically, it would interesting to investigate if SFDA via self-supervised learning could be extended to other downstream computer vision problems such as object recognition, image classification, scene understanding, tracking, and so on.
Moreover, it would interesting to investigate if SFDA via self-supervised learning could be extended to other  computer vision problems such as object recognition, image classification, scene understanding, etc.

{\small
\bibliographystyle{ieee_fullname}
\bibliography{references}
}

\end{document}